\documentclass[nohyperref]{article}

\usepackage{microtype}
\usepackage{graphicx}
\usepackage{subfigure}
\usepackage{booktabs} % for professional tables

\usepackage{hyperref}

\usepackage[accepted]{icml2022}

\usepackage{hyperref}

\usepackage{bbold}
\usepackage{bm}
\usepackage{amsmath,amscd,amssymb}
\usepackage{xcolor}
\usepackage{colortbl}
\usepackage{tabu}
\usepackage{thm-restate,cleveref}
\usepackage{wrapfig,caption}

\newcommand{\post}{{\rm post}}

\newcommand{\dais}{{\rm DAIS}}
\newcommand{\KL}{{\rm KL}}
\newcommand{\agg}{{\rm Agg}}
\newcommand{\sldais}{{\rm SL-DAIS}}
\newcommand{\nsdais}{{\rm NS-DAIS}}

\newcommand{\ty}{\tilde{y}}

\newcommand{\tbx}{\mathbf{\tilde{x}}}
\newcommand{\by}{\mathbf{y}}
\newcommand{\bw}{\mathbf{w}}
\newcommand{\bW}{\mathbf{W}}
\newcommand{\tbz}{\tilde{\mathbf{z}}}

\newcommand{\tbW}{\tilde{\mathbf{W}}}
\newcommand{\bA}{\mathbf{a}}
\newcommand{\dbA}{\delta\mathbf{a}}
\newcommand{\bB}{\mathbf{B}}
\newcommand{\dbB}{\delta\mathbf{B}}
\newcommand{\bZ}{\mathbf{Z}}
\newcommand{\bmu}{\bm{\mu}}
\newcommand{\tbmu}{\bm{\tilde{\mu}}}
\newcommand{\sigmaobs}{\sigma_{\rm obs}}
\newcommand{\bLam}{\bm{\Lambda}}
\newcommand{\tbLam}{\bm{\tilde{\Lambda}}}

\newcommand{\bx}{\mathbf{x}}
\newcommand{\bX}{\mathbf{X}}
\newcommand{\bz}{\mathbf{z}}
\newcommand{\bv}{\mathbf{v}}
\newcommand{\bg}{\mathbf{g}}
\newcommand{\bu}{\mathbf{u}}
\newcommand{\bff}{\mathbf{f}}
\newcommand{\bK}{\mathbf{K}}
\newcommand{\bk}{\mathbf{k}}
\newcommand{\Nsurr}{N_{\rm surr}}
\newcommand{\zero}{\mathbf{0}}
\newcommand{\NN}{\mathcal{N}}
\newcommand{\II}{\mathcal{I}}
\newcommand{\JJ}{\mathcal{J}}
\newcommand{\RR}{\mathbb{R}}
\newcommand{\mass}{\mathbf{M}}
\newcommand{\eps}{{\bm{\varepsilon}}}
\newcommand{\OO}{\mathcal{O}}
\newcommand{\LL}{\mathcal{L}}
\newcommand{\TT}{\mathcal{T}}
\newcommand{\tTT}{\tilde{\mathcal{T}}}
\newcommand{\EE}{\mathbb{E}}
\newcommand{\qf}{q_{\rm fwd}}
\newcommand{\qb}{q_{\rm bwd}}

\newcommand{\tqb}{\tilde{q}_{\rm bwd}}
\newcommand{\DD}{\mathcal{D}}
\newcommand{\PsiL}{\Psi_{\rm L}}
\newcommand{\PsiLhat}{\hat{\Psi}_{\rm L}}
\newcommand{\pth}{p_{\theta}}
\newcommand{\bomega}{\bm{\omega}}
\newcommand{\trace}{\mathrm{Tr}}

\icmltitlerunning{Surrogate Likelihoods for Variational Annealed Importance Sampling}

\begin{document}

\twocolumn[
\icmltitle{Surrogate Likelihoods for Variational Annealed Importance Sampling}




\begin{icmlauthorlist}
\icmlauthor{Martin Jankowiak}{broad}
\icmlauthor{Du Phan}{google}
\end{icmlauthorlist}

\icmlaffiliation{broad}{Broad Institute, Cambridge, MA, USA}
\icmlaffiliation{google}{Google Research, Cambridge, MA, USA}

\icmlcorrespondingauthor{Martin Jankowiak}{mjankowi@broadinstitute.org}

\icmlkeywords{Variational inference, Approximate bayesian inference, Hamltonian Monte Carlo, Annealed Importance Sampling}

\vskip 0.3in
]


\printAffiliationsAndNotice{}  % leave blank if no need to mention equal contribution

%%%%%%%%%%%%%%%%%%%%%%%%%%%%%%%%%%%%%%%%%%%
\begin{abstract}
Variational inference is a powerful paradigm for approximate Bayesian inference with a number of appealing 
properties, including support for model learning and data subsampling. By contrast MCMC methods like Hamiltonian
Monte Carlo do not share these properties but remain attractive since, contrary to parametric methods,
MCMC is asymptotically unbiased. For these reasons researchers have sought to combine the strengths of both
classes of algorithms, with recent approaches coming closer to realizing this vision in practice. 
However, supporting data subsampling in these hybrid methods can be a challenge, a shortcoming that we address by introducing 
a surrogate likelihood that can be learned jointly with other variational parameters.
We argue theoretically that the resulting algorithm allows an intuitive trade-off between
inference fidelity and computational cost.
In an extensive empirical comparison we show that our method performs well in practice and that it is
well-suited for black-box inference in probabilistic programming frameworks.
\end{abstract}
%%%%%%%%%%%%%%%%%%%%%%%%%%%%%%%%%%%%%%%%%%%

%%%%%%%%%%%%%%%%%%%%%%%%%%%%%%%%%%%%%%%%%%%
\section{Introduction}
\label{sec:intro}
%%%%%%%%%%%%%%%%%%%%%%%%%%%%%%%%%%%%%%%%%%%

Bayesian modeling and inference is a powerful approach to making sense of complex datasets.
This is especially the case in scientific applications, where accounting for prior knowledge
and uncertainty is essential.
For motivating examples we need only look to recent efforts to study COVID-19, including
e.g.~epidemiological models that incorporate mobility data \citep{miller2020mobility,monod2021age}
or link differentiable transmissibility of SARS-CoV-2 lineages to viral mutations
\citep{obermeyer2021analysis}.

Unfortunately for many application areas the broader use of Bayesian models is hampered by 
the difficulty of formulating scalable inference algorithms.
One regime that has proven particularly challenging is the large data regime.
This is essentially because powerful MCMC methods like Hamiltonian Monte Carlo (HMC) \citep{duane1987hybrid, neal2011mcmc}
are, at least on the face of it, incompatible with data subsampling (i.e.~mini-batching) \citep{betancourt2015fundamental}.
While considerable effort has gone into developing MCMC methods that accommodate
data subsampling,\footnote{See e.g.~\citet{dang2019hamiltonian} and \citet{zhang2020asymptotically} for recent work.} 
developing generic MCMC methods that yield high fidelity posterior approximations 
while scaling to very large datasets remains challenging.

For this reason, among others, recent years have seen extensive development of approximate inference methods based 
on variational inference \citep{jordan1999introduction,blei2017variational}.
Besides `automatic' support for data subsampling \citep{hoffman2013stochastic,ranganath2014black}, at least for suitable model classes, variational inference 
has several additional favorable properties, including support for amortization \citep{dayan1995helmholtz},
log evidence estimates, and model learning, motivating researchers to combine
the strengths of MCMC and variational inference \citep{salimans2015markov,hoffman2017learning,caterini2018hamiltonian,ruiz2019contrastive}.

Recent work from \citet{geffner2021mcmc} and \citet{zhang2021differentiable} offers a particularly elegant formulation 
of a variational inference method that leverages (unadjusted) HMC as well as annealed importance sampling (AIS) \citep{neal2001annealed}.
Although these are fundamentally variational methods, since they make use of HMC, which does not itself readily accommodate data subsampling,
these methods likewise do not automatically inherit support for data subsampling. 
This is unfortunate because these methods \emph{do} automatically inherit many
of the other nice features of variational inference, including support for amortization and model learning. 

In this work we set out to extend the approach in \citet{geffner2021mcmc} and \citet{zhang2021differentiable} 
to support data subsampling, thus making it applicable to the large data regime. 
Our basic strategy is simple and revolves around introducing a \emph{surrogate log likelihood} 
that is cheap to evaluate. The surrogate log likelihood is used to guide HMC dynamics, resulting in a flexible variational distribution
that is implicitly defined via the gradient of the surrogate log likelihood. The corresponding variational objective
accommodates unbiased mini-batch estimates, at least for the large class of models with appropriate conditional independence structure that we consider.
As we show in experiments in Sec.~\ref{sec:exp}, our method performs well in practice and is
well-suited for black-box inference in probabilistic programming frameworks.

%%%%%%%%%%%%%%%%%%%%%%%%%%%%%%%%%%%%%%%%%%%
\section{Problem setting}
\label{sec:setting}
%%%%%%%%%%%%%%%%%%%%%%%%%%%%%%%%%%%%%%%%%%%

We are given a dataset $\DD$ and a model of the form
%%%
\vspace{-1.5mm}
\begin{align}
\label{eqn:model}
\pth(\DD, \bz) = \pth(\bz) \prod_{n=1}^N \pth(y_n | \bz, \bx_n)
\end{align}
%%%
where the latent variable $\bz \in \RR^D$ is governed
by a prior $\pth(\bz)$ and the likelihood factorizes
into $N$ terms, one for each data point $(y_n, \bx_n) \in \DD$,
and where $\theta$ denotes any additional (non-random) parameters in the model.
Eqn.~\ref{eqn:model} encompasses a large and important class of models
and includes e.g.~a wide variety of multi-level regression models. 
The log density corresponding to Eqn.~\ref{eqn:model} is given by
%%%
\begin{align}
\begin{split}
    \log \pth(\DD, \bz) &= \log \pth(\bz) + \Sigma_n \log \pth(y_n | \bz, \bx_n) \\ 
    &\equiv \Psi_0(\bz) + \PsiL(\DD, \bz)
    %&\equiv \Psi(\bz) \equiv \Psi_0(\bz) + \PsiL(\DD, \bz)
\end{split}
\end{align}
%%%
where we define % the full log density $\Psi(\bz)$
the log prior $\Psi_0(\bz)$ and 
log likelihood $\PsiL(\DD, \bz)$.\footnote{We also use the notation 
$\PsiL(\DD_\II, \bz) \equiv \Sigma_{n \in \II} \log \pth(y_n | \bz, \bx_n)$ for a set of indices $\II$.}

We are interested in the regime where $N$ is large or individual likelihood terms are costly to
evaluate so that computing the full log likelihood $\PsiL(\DD, \bz)$ and its gradients is impractical. 
We aim to devise a flexible variational approximation to the 
posterior $\pth(\bz | \DD)$ that can be fit with an algorithm that supports data subsampling.
Additionally we would like our method to be generic in nature so that it can readily be
incorporated into a probabilistic programming framework as a black-box inference algorithm.
Finally we would like a method that supports model learning, i.e.~one that allows us to learn
$\theta$ in conjunction with the approximate posterior.

For simplicity in this work we primarily focus on global latent variable models with the structure in Eqn.~\ref{eqn:model}.
The approach we describe can also be extended to models with \emph{local} latent variables,
i.e.~those local to each data point.
For more discussion see Sec.~\ref{sec:llv} and Sec.~\ref{app:llvm} in the appendix.

%%%%%%%%%%%%%%%%%%%%%%%%%%%%%%%%%%%%%%%%%%%
\section{Background}
\label{sec:bg}
%%%%%%%%%%%%%%%%%%%%%%%%%%%%%%%%%%%%%%%%%%%

Before describing our method in Sec.~\ref{sec:method}, we first review relevant background.

%%%%%
\subsection{Variational inference}
%%%%%

The simplest variants of variational inference introduce a parametric
variational distribution $q_\phi(\bz)$ and proceed to choose the parameters $\phi$
to minimize the Kullback-Leibler (KL) divergence between the variational distribution and the posterior $\pth(\bz | \DD)$, 
i.e.~${\rm KL}( q_\phi(\bz) | \pth(\bz | \DD))$. 
This can be done by maximizing the Evidence Lower Bound or ELBO
%%%
\begin{equation}
    \label{eqn:elbo}
    {\rm ELBO} \equiv \EE_{q_\phi(\bz)} \left[ \log \pth(\DD, \bz) - \log q_\phi(\bz)\right]
\end{equation}
%%%
which satisfies ${\rm ELBO} \le \log \pth(\DD)$. Thanks to this inequality 
the ELBO naturally enables joint model learning and inference, i.e.~we can 
maximize Eqn.~\ref{eqn:elbo} w.r.t.~both model parameters $\theta$ and variational parameters $\phi$ simultaneously.
A potential shortcoming of the fully parametric approach described here is that it can be difficult
to specify appropriate parameterizations for $q_\phi(\bz)$.

%%%%%
\subsection{Annealed Importance Sampling}
%%%%%

Annealed importance sampling (AIS) \citep{neal2001annealed} is a method for estimating the 
evidence $\pth(\DD)$ that leverages a sequence of $K$ bridging densities $\{f_k(\bz)\}_{k=1}^K$ that connect
a simple base distribution $q_0(\bz)$ to the posterior. In more detail, AIS can be understood as importance sampling on an extended space.
That is we can write 
%%%
\begin{equation}
\begin{split}
    \label{eqn:ais}
%\pth(\DD) &= \int\! d\bz \; \pth(\DD, \bz)  \\
%&= \EE_{\qf(\bz_{0:{K}})} \left[ \qb(\bz_{0:{K}}) /  \qf(\bz_{0:{K}}) \right] 
\pth(\DD) \!=\! \int\! d\bz \; \pth(\DD, \bz)  
    = \EE_{\qf(\bz_{0:{K}})} \left[ \frac{\qb(\bz_{0:{K}}) }{ \qf(\bz_{0:{K}}) } \right] 
\end{split}
\end{equation}
%%%
where the proposal distribution $\qf$ and the (un-normalized) target distribution $\qb$ are given by
%%%
\begin{equation}
\begin{split}
    \label{eqn:qfqb}
\qf(\bz_{0:{K}}) &= q_0(\bz_0) \TT_1(\bz_1 | \bz_0) \cdots \TT_{K}(\bz_{K} | \bz_{K-1}) \\
\qb(\bz_{0:{K}}) &= \pth(\DD, \bz_K) \tTT_{K}(\bz_{K-1} | \bz_{K}) \cdots \tTT_1(\bz_0 | \bz_{1}) \nonumber
\end{split}
\end{equation}
%%%
Here each $\TT_k$ is a MCMC kernel that leaves the bridging density $f_k(\bz)$ invariant.
While there is considerable freedom in the choice of $q_0(\bz)$ and $\{f_k(\bz)\}$, 
it is natural to let $q_0(\bz)=\pth(\bz)$ and $f_k(\bz) \propto q_0(\bz)^{1-\beta_k} \pth(\DD, \bz)^{\beta_k}$ 
where $\{ \beta_k\}$ are inverse temperatures that satisfy $0 < \beta_1 < \beta_2 < ... < \beta_K = 1$.
Additionally
%%%
\begin{equation}
\tTT_k(\bz_{k-1} | \bz_k) = \TT_k(\bz_k | \bz_{k-1})f_k(\bz_{k-1})/f_k(\bz_k)
\end{equation}
%%%
is the reverse MCMC kernel corresponding to $\TT_k$.
Conceptually, the kernels $\{\TT_k\}$ move samples from $q_0(\bz)$ towards the posterior
via a sequence of moves, each of which is `small' for appropriately spaced $\{\beta_k\}$ and sufficiently large $K$.
Indeed AIS is consistent as $K \rightarrow \infty$ \citep{neal2001annealed} and has been shown
to achieve accurate estimates of $\log \pth(\DD)$ empirically \citep{grosse2015sandwiching}.

%%%%%
\subsection{Hamiltonian Monte Carlo}
%%%%%

Hamiltonian Monte Carlo (HMC) is a powerful gradient-based MCMC method \citep{duane1987hybrid, neal2011mcmc}.
HMC proceeds by introducing an auxiliary momentum $\bv \in \RR^D$ and a log joint (un-normalized) density $\log \pi(\bz, \bv) = -H(\bz, \bv)$
where the Hamiltonian is defined as $H(\bz, \bv) = T(\bv) + V(\bz)$, $V(\bz) = -\log \pth(\DD, \bz)$ is the
potential energy, $T(\bv) = \tfrac{1}{2} \bv^\top \mass^{-1} \bv$ is the kinetic energy, and $\mass$ is the mass matrix.
Hamiltonian dynamics is then defined as:
%%%
\begin{equation}
    \label{eqn:hamilton}
    \frac{d\bz}{dt} = \frac{\partial H}{\partial \bv} \qquad
    \frac{d\bv}{dt} = -\frac{\partial H}{\partial \bz}
\end{equation}
%%%
Evolving trajectories according to Eqn.~\ref{eqn:hamilton} for any time $\tau$
yields Markovian transitions that target the stationary distribution $\pi(\bz, \bv)$. 
By alternating Hamiltonian evolution with momentum updates
%%%
\begin{equation}
    \bv_{t+1} \sim \NN(\gamma \bv_t, (1-\gamma^2) \mass)
\end{equation}
%%%
and disregarding the momentum yields a MCMC chain targeting the posterior $\pth(\bz | \DD)$.
Here $\gamma \in [0, 1)$ controls the level of momentum refreshment; e.g.~the $\gamma \approx 1$ regime suppresses random-walk behavior \citep{horowitz1991generalized}. 
In general we cannot simulate Hamiltonian trajectories exactly and instead use a symplectic integrator like the so-called `leapfrog' integrator.
Combined with momentum refreshment a single HMC step $(\bz_{k-1}, \bv_{k-1}) \rightarrow (\bz_k, \bv_k)$
is given by
%%%
\begin{align}
    \label{eqn:leap}
    &\hat{\bz}_k \!\leftarrow\! \bz_{k-1} \!+\! \tfrac{\eta}{2} \mass^{-1} \bv_{k-1} \;
    &&\hat{\bv}_k \!\leftarrow\! \bv_{k-1} \!-\! \eta \nabla V(\hat{\bz}_k)  \\
    &\bz_k \!\leftarrow\!\hat{\bz}_k \!+\! \tfrac{\eta}{2} \mass^{-1} \hat{\bv}_k  \;
    &&\bv_k \!\sim\! \NN(\gamma \hat{\bv}_k, (1\!-\!\gamma^2) \mass) \nonumber
\end{align}
%%%
where $\eta$ is the step size. Since the leapfrog integrator is inexact, a Metropolis-Hastings accept/reject step is used
to ensure asymptotic correctness.
For a comprehensive introduction to HMC see e.g.~\citep{betancourt2017conceptual}.

%%%%%
\subsection{Differentiable Annealed Importance Sampling a.k.a.~Uncorrected Hamiltonian Annealing}
%%%%%

We describe recent work, (UHA; \citet{geffner2021mcmc}) and (DAIS; \citet{zhang2021differentiable}),
that combines HMC and AIS in a variational inference framework.
For simplicity we refer to the authors' algorithm as DAIS and disregard any differences
between the two (contemporaneous) references.\footnote{A closely related approach that utilizes
unadjusted overdamped Langevin steps is described in \cite{thin2021monte}.}

A notable feature of HMC is that the HMC kernel $\TT(\bz_1, \bv_1 | \bz_0, \bv_0)$ can generate large
moves in $\bz$-space with large acceptance probabilities, which makes it an attractive kernel choice for AIS
\citep{sohl2012hamiltonian}.
Moreover, as an importance sampling framework AIS naturally gives rise to a variational bound, since
applying Jensen's inequality to the log of Eqn.~\ref{eqn:ais} immediately yields 
a lower bound to $\log p(\DD)$.
\citet{geffner2021mcmc} and \citet{zhang2021differentiable} note that the utility of such an AIS variational
bound is severely hampered by the fact that typical MCMC kernels include a Metropolis-Hastings accept/reject
step that makes the bound non-differentiable.
DAIS restores differentiability by removing the accept/reject step,
which makes it straightforward to optimize the variational bound using gradient-based methods.
While dropping the Metropolis-Hastings correction invalidates detailed balance,
AIS and thus the resulting variational bound remain intact. 
Moreover, since DAIS employs a HMC kernel and since we expect HMC moves to have high acceptance 
probabilities---at least for a well-chosen step size $\eta$ and mass matrix $\mass$---we
expect the $K$ DAIS transitions to efficiently move samples from $q_0(\bz)$ towards the posterior.

In more detail DAIS operates on an extended space $(\bz_0, ..., \bz_K, \bv_0, ..., \bv_K)$,
with proposal and target distributions given by
%%%
\begin{align}
\label{eqn:qfqbdais}
    \qf(\bz_{0:K}, \bv_{0:K}) &= q_0(\bz_0) q_0(\bv_0)  \times \\
    &\textstyle{\prod}_{k=1}^K \TT_k(\bz_k, \bv_k | \bz_{k-1}, \bv_{k-1}) \nonumber \\
\qb(\bz_{0:K}, \bv_{0:K}) &= \pth(\DD, \bz_K) \times \\
    &\textstyle{\prod}_{k=1}^K \tTT_k(\bz_{k-1}, \bv_{k-1} | \bz_k, \bv_k) \nonumber
\end{align}
%%%
where $q_0(\bv_0)=\NN(\bv_0 | \zero, \mass)$ is the momentum distribution.
Here each kernel $\TT_k$ performs a single leapfrog step as in Eqn.~\ref{eqn:leap} using the annealed potential energy 
%%%
\begin{align}
    \label{eqn:annealedV}
V_k(\bz) &= -(1-\beta_k) \log q_0(\bz) - \beta_k \log \pth(\DD, \bz)  \\
         &= -(1-\beta_k) \log q_0(\bz) - \beta_k \left( \Psi_0(\bz) + \PsiL(\DD, \bz) \right)  \nonumber
\end{align}
%%%
and without including a Metropolis-Hastings correction. 
Notably, \citet{geffner2021mcmc} and \citet{zhang2021differentiable} show
that the DAIS variational objective is easy to compute, as clarified by the following lemma.
%%%
\begin{restatable}{lemma}{lemmaone}
\label{lemma:one}
%%%
The DAIS bound given by proposal and target distributions as in Eqn.~\ref{eqn:qfqbdais} 
is differentiable and is given by 
%%%
\begin{equation}
\begin{split}
    \nonumber
    \LL_\dais &\equiv \EE \Big[ \log \qb(\bz_{0:K}, \bv_{0:K}) \!-\! \log  \qf(\bz_{0:K}, \bv_{0:K}) \Big]  \\
    &= \EE \Big[ \log \pth(\DD, \bz_K) - \log q_0(\bz_0) + \\
                       &\;\;\qquad\textstyle{\sum}_{k=1}^K  \left \{ \log \NN(\hat{\bv}_k, \mass) - \log\NN(\bv_{k-1}, \mass) \right \} \Big]
\end{split}
\end{equation}
%%%
where the expectation is w.r.t.~$\qf$.
Moreover, gradient estimates of $\LL_\dais$ w.r.t.~$\theta$ and $\phi$ can be computed using reparameterized gradients
    provided the base distribution $q_0(\bz)$ is reparameterizable.
For a proof see Sec.~\ref{app:daiselbo} in the supplemental materials, \citet{geffner2021mcmc} or \citet{zhang2021differentiable}.
%%%
\end{restatable}
%%%
Note that $\TT_k$ and $\tTT_k$ do not appear explicitly in Lemma~\ref{lemma:one}; instead
it suffices to compute kinetic energy differences.

%%%%%%%%%%%%%%%%%%%%%%%%%%%%%%%%%%%%%%%%%%%
\section{Surrogate Likelihood DAIS}
\label{sec:method}
%%%%%%%%%%%%%%%%%%%%%%%%%%%%%%%%%%%%%%%%%%%

The DAIS variational objective in Lemma~\ref{lemma:one} can lead to tight bounds on the
model log evidence $\log \pth(\DD)$, especially for large $K$. However, for a
model like that in Eqn.~\ref{eqn:model}, optimizing $\LL_\dais$ can be prohibitively 
expensive, with a $\OO(NK)$ cost per optimization step for a dataset with $N$ data points.
This is because sampling $\qf$ requires computing the gradient of the full log likelihood,
i.e.~$\nabla \PsiL(\DD, \bz)$, $K$ times, which is expensive whenever $N$ is large or individual
likelihood terms are costly to compute.

In order to speed-up training and sampling in this regime we first observe that the
proof of Lemma~\ref{lemma:one} holds for \emph{any choice} of potential energies $\{V_k(\bz)\}$.\footnote{See
Sec.~\ref{app:daiselbo} in the supplement for details.}
The annealed ansatz in Eqn.~\ref{eqn:annealedV} is a natural choice, but any
other choice that efficiently moves samples from $q_0(\bz)$ towards the posterior $\pth(\bz|\DD)$ can lead to tight variational bounds.
This observation naturally leads to two scalable variational bounds that are tractable in the large data regime.
In the first method, a variant of which was considered theoretically in \citet{zhang2021differentiable} and 
which we refer to as {\bf NS-DAIS}, we replace $V_k(\bz)$ with a \emph{stochastic} estimate 
that depends on a mini-batch of data. 
In the second method, which we refer to as {\bf SL-DAIS},
we replace the full log likelihood $\PsiL(\DD, \bz)$ with a \emph{surrogate log likelihood}
$\PsiLhat(\bz)$ that is cheap to evaluate.  Below we argue theoretically and demonstrate empirically
that SL-DAIS is superior to NS-DAIS.

%%%%%%%%%%%%%%%%%%%%
\subsection{NS-DAIS: Naive Subsampling DAIS}
\label{sec:nsdais}
%%%%%%%%%%%%%%%%%%%%

For any mini-batch of indices $\JJ \subset \{1, ..., N\}$ with $|\JJ| = B$
we can define an estimator of $\PsiL(\DD, \bz)$ as $\frac{N}{B}\PsiL(\DD_\JJ, \bz)$.
In NS-DAIS we plug this estimator into the potential energy in Eqn.~\ref{eqn:annealedV}.
With this choice of $V_k(\bz)$ sampling from $\qf$ is $\OO(B)$ instead of $\OO(N)$. 
Moreover, by replacing the $\log \pth(\DD, \bz_K)$ term in $\LL_\dais$ with an estimator
computed using an \emph{independent} mini-batch of indices $\II \subset \{1, ..., N\}$ with
$|\II| = B$ we obtain an unbiased estimator that is $\OO(B)$ instead
of $\OO(N)$. 
More formally, the proposal and target distributions $\qf$ and $\qb$
in Eqn.~\ref{eqn:qfqbdais} are replaced with
%%%
\begin{equation}
\begin{split}
    \label{eqn:qfqbnsdais}
    &\qf \rightarrow q_0(\bz_0) q_0(\bv_0) \textstyle{\prod}_{k=1}^K \TT_k(\bz_k, \bv_k | \bz_{k-1}, \bv_{k-1}, \JJ) \\
    &\qb \rightarrow \pth( \bz_K | \DD_{\II})^{N/B} \pth(\bz_K) \textstyle{\prod}_{k=1}^K \tTT_k(\bz_{k-1}, \bv_{k-1} | \cdot) \nonumber
    %&\qb \rightarrow \pth( \bz_K | \DD_{\II})^{N/B} \pth(\bz_K) \textstyle{\prod}_{k=1}^K \tTT_k(\bz_{k-1}, \bv_{k-1} | \bz_k, \bv_k, \JJ) \nonumber
\end{split}
\end{equation}
%%%
where we omit the common factor of $q(\II) q(\JJ)$ for brevity. 
See Algorithm~\ref{algo:nsdais} and Sec.~\ref{app:nsdais} in the supplement for details.

%%%%%%%%%%%%%%%%%%%%
\subsection{SL-DAIS: Surrogate Likelihood DAIS}
\label{sec:sldais}
%%%%%%%%%%%%%%%%%%%%

\definecolor{MyBlue}{RGB}{176,196,222}

\newcommand{\highlight}[1]{%
  \colorbox{MyBlue!70}{$\displaystyle#1$}}

\begin{algorithm}[tb]
    \caption{SL-DAIS: Surrogate Likelihood Differentiable Annealed Importance Sampling. We
    highlight in \highlight{\textrm{blue}} where the algorithm differs from DAIS. 
    To recover DAIS we substitute $\PsiLhat(\bz) \!\rightarrow\! \PsiL(\DD, \bz)$ and $B \!\rightarrow\! N$. 
    Note that $\PsiLhat$ is only used to guide HMC dynamics and that a stochastic estimate of $\PsiL(\DD, \bz)$ still appears 
    on the final line. %See Algorithm~\ref{algo:dais} in the supplement for DAIS.
    }
   \label{algo:sldais}
%%%%%%%%%%%%%%%
\begin{algorithmic}
    \STATE {\bfseries Input:} model log density $\Psi_0(\bz) + \PsiL(\DD, \bz)$, 
                              surrogate log likelihood $\PsiLhat(\bz)$,
                              base variational distribution $q_0(\bz)$, number of steps $K$,
                              inverse temperatures $\{ \beta_k \}$, 
                              step size $\eta$, momentum refresh parameter $\gamma$, 
                              mass matrix $\mass$,
                              dataset $\DD$ of size $N$,
                              mini-batch size $B$
    \STATE {\bfseries Initialize:}  $\bz_0 \sim q_0$, $\bv_0 \sim \NN(\zero, \mass)$, $\LL \leftarrow -\log q_0 (\bz_0)$
    %%%%
    \FOR{$k=1$ {\bfseries to} $K$}
    %%%
    \STATE $\hat{\bz}_k \leftarrow \bz_{k-1} + \frac{\eta}{2}\mass^{-1}\bv_{k-1}$ 
    \STATE $\bg_k \leftarrow \! \nabla_{\!\bz} \! \left\{\!\beta_k \!\left( \! \Psi_0(\bz) \!+\! 
                                               \highlight{\PsiLhat(\bz)} \! \right) \!+\! 
                                               (1 \!-\! \beta_k) \log q_0(\bz)\right\}\!\bigg\rvert_{\bz=\hat{\bz}_k} $
    \STATE $\hat{\bv}_k \leftarrow \bv_{k-1} + \eta \bg_k $
    \STATE $\bz_{k} \leftarrow \hat{\bz}_k+ \frac{\eta}{2}\mass^{-1}\hat{\bv}_{k}$
    \vspace{1mm}
    \IF{$k < K$}
        \STATE $\bv_k \leftarrow \gamma \hat{\bv}_k + \sqrt{1 - \gamma^2}\eps, \; \eps \sim \NN(\zero, \mass)$
    \ENDIF
    \STATE $\LL \leftarrow \LL + \log \NN(\hat{\bv}_k, \mass) - \log \NN(\bv_{k-1}, \mass)$
    %%%
    \ENDFOR \\
    \highlight{\textrm{Sample}} mini-batch indices $\mathcal{I} \! \subset \! \{1, ..., N\}$ with $|\mathcal{I}|=B$
    \STATE {\bfseries Return:} $\LL \mathrel{+} \Psi_0(\bz_K) + \highlight{\tfrac{N}{B} \PsiL(\DD_{\mathcal{I}}, \bz_K)} $ 
    %%%
\end{algorithmic}
\end{algorithm}

We proceed as in NS-DAIS except we plug in
a \emph{fixed non-stochastic} surrogate log likelihood
$\PsiLhat(\bz)$ into the potential energy in Eqn.~\ref{eqn:annealedV}.
More formally, the proposal and target distributions $\qf$ and $\qb$
in Eqn.~\ref{eqn:qfqbdais} are replaced with
%%%
\begin{equation}
\begin{split}
    \label{eqn:qfqbsldais}
    &\qf \rightarrow  q(\II)q_0(\bz_0)q_0(\bv_0) \textstyle{\prod}_{k=1}^K \TT_k(\bz_k, \bv_k | \bz_{k-1}, \bv_{k-1}, \PsiLhat) \\
    &\qb \rightarrow q(\II)\pth(\DD_{\II} | \bz_K)^{N/B} \pth(\bz_K) \textstyle{\prod}_{k=1}^K \tTT_k(\bz_{k-1}, \bv_{k-1} | \cdot) \nonumber
    %&\qb \rightarrow q(\II)\pth(\DD_{\II} | \bz_K)^{N/B} \pth(\bz_K) \textstyle{\prod}_{k=1}^K \tTT_k(\bz_{k-1}, \bv_{k-1} | \bz_k, \bv_k, \PsiLhat) \nonumber
\end{split}
\end{equation}
%%%
where $q(\II)$ encodes sampling mini-batches of $B$ indices without replacement,
and data subsampling occurs solely in the $\pth(\DD_{\II} | \bz_K)$ term.
Like NS-DAIS the SL-DAIS ELBO admits a simple unbiased gradient estimator.
In contrast to NS-DAIS, the HMC dynamics in SL-DAIS targets a fixed target distribution.
We expect the reduced stochasticity of SL-DAIS as compared to NS-DAIS to result in superior empirical
performance.\footnote{This expectation is closely related to a comment in \citet{zhang2021differentiable}: \emph{[DAIS] relies on all intermediate distributions, and so the error induced by stochastic gradient noise accumulates over the whole trajectory. Simply taking smaller steps
fails to reduce the error. We conjecture that AIS-style algorithms are inherently fragile to gradient noise.}}
See Algorithm~\ref{algo:sldais} for the complete algorithm and Sec.~\ref{app:sldais} in the supplement for
a more complete formal description.

There are multiple possibilities for how to parameterize $\PsiLhat(\bz)$.
We briefly describe the simplest possible recipe, which we refer to as \texttt{RAND},
leaving a detailed ablation study to Sec.~\ref{sec:ablation}.
In \texttt{RAND} we randomly choose $\Nsurr \ll N$ surrogate data points $\{ (\ty_n, \tbx_n) \} \subset \DD$
and introduce a $\Nsurr$-dimensional vector of learnable (positive) weights $\bomega$, 
where $\bomega$ can be learned jointly with other variational parameters.
The surrogate log likelihood is then given by
%%%
\begin{equation}
\label{eqn:surrogateansatz}
\PsiLhat(\bz) = \sum_n \omega_n \log p_\theta(\ty_n | \bz, \tbx_n)
\end{equation}
%%%
Evidently computing $\PsiLhat(\bz)$ is $\OO(\Nsurr)$.
Besides its simplicity, an appealing feature of this parameterization 
is that it leverages the known functional form of the likelihood.\footnote{\citet{chen2022bayesian} use the same ansatz in concurrent work focused on Bayesian coresets.}

%%%%%%%%%%%%%%%%%%%%
\subsection{Discussion}
\label{sec:methdisc}
%%%%%%%%%%%%%%%%%%%%

Before taking a closer look at the theoretical properties of NS-DAIS and SL-DAIS in Sec.~\ref{sec:theory},
we make two simple observations. First, as we should expect from any bona fide variational method,  
maximizing the variational bound does indeed lead to tighter posterior approximations, as formalized
in the following proposition:
%%%
\begin{restatable}{prop}{propnssl}
\label{prop:nssl}
The NS-DAIS and SL-DAIS approximate posterior distributions, each of which is given by the marginal $\qf(\bz_K)$, both satisfy the inequality
%%%
\begin{equation}
\begin{split}
    \nonumber
    \log \pth(\DD) - \LL  \ge \KL( \qf(\bz_{K}) | \pth(\bz_K | \DD) )  \ge 0
\end{split}
\end{equation}
%%%
where $\LL$ is $\LL_\nsdais$ or $\LL_\sldais$, respectively.
\end{restatable}
%%%
See Sec.~\ref{app:nssldais} for a proof.
Thus as $\LL$ increases for fixed $\theta$, the KL divergence decreases and $\qf(\bz_{K})$ becomes a better approximation to the posterior.
Second, sampling from $\qf$ in the case of NS-DAIS requires the entire dataset $\DD$. Conveniently in the case of SL-DAIS we
only require the surrogate log likelihood $\PsiLhat(\bz)$ so that the dataset can be discarded after training.

%%%%%%%%%%%%%%%%%%%%%%%%%%%%%%%%%%%%%%%%%%%
\section{Convergence analysis}
\label{sec:theory}
%%%%%%%%%%%%%%%%%%%%%%%%%%%%%%%%%%%%%%%%%%%

As in \citet{zhang2021differentiable} to make our analysis tractable we consider linear regression
with a prior $\pth(\bz) = \NN(\bmu_0, \bLam_0^{-1})$ and a likelihood $\prod_n \NN(y_n | \bz \cdot \bx_n, \sigmaobs^2)$,
where each $y_n \in \RR$ and $\bx_n \in \RR^D$.
Furthermore we work under the following set of simplifying assumptions:
%%%
\vspace{-2mm}
\begin{restatable}{ass}{assumption}
\label{ass}
%We use full momentum refreshment ($\gamma = 0$), 
We use $\gamma=0$, equally spaced inverse temperatures $\{ \beta_k \}$,
a step size that varies as $\eta \sim K^{-1/4}$,
and the prior as the base distribution (i.e.~$q_0(\bz) \!=\! \pth(\bz)$).
\end{restatable}
\vspace{-3mm}

%%%%%%%%%%%%%%%%%%%%
\subsection{NS-DAIS}
\label{sec:nldaistheory}
%%%%%%%%%%%%%%%%%%%%

What kind of posterior approximation do we expect from NS-DAIS? As clarified
by the following proposition, NS-DAIS does not directly target the posterior $\pth(\bz | \DD)$.
%%%
\begin{restatable}{prop}{propnsdais}
\label{prop:nsdais}
The approximate posterior for linear regression 
given by running NS-DAIS with $K$ steps under Assumption~\ref{ass} 
converges to the `aggregate pseudo-posterior'
%%%
\begin{equation}
     \pth^\agg(\bz | \DD) \equiv \EE_{q(\JJ)} \left[ \pth(\bz | \DD_\JJ) \right]
\end{equation}
%%%
as $K \rightarrow \infty$. Here $\pth(\bz | \DD_\JJ)$ denotes the posterior corresponding
to the data subset $\DD_\JJ$ with appropriately scaled likelihood term, 
i.e.~$\pth(\bz | \DD_\JJ) \propto \pth(\DD_\JJ | \bz)^{N/B} \pth(\bz)$.
%%%
\end{restatable}
%%%
See Sec.~\ref{app:nsdais} for a proof and additional details.
Thus we generally expect NS-DAIS to provide a good posterior approximation
when the aggregate pseudo-posterior is a good approximation to the posterior.
The poor performance of NS-DAIS in experiments in Sec.~\ref{sec:exp} suggests 
that this is not case for moderate mini-batch sizes in typical models.
%Indeed this result is not surprising, as \citet{zhang2021differentiable} prove\footnote{In the case of Bayesian
%linear regression and assuming $\gamma=0$; see Theorem 2 in \citet{zhang2021differentiable} for details.}
%that NS-DAIS is inconsistent, in the sense that $\LL_\nsdais$ does not converge
%to $\log \pth (\DD)$ as $K\rightarrow \infty$.

%%%%%%%%%%%%%%%%%%%%
\subsection{SL-DAIS}
\label{sec:sldaistheory}
%%%%%%%%%%%%%%%%%%%%

We now analyze SL-DAIS assuming the surrogate log likelihood $\PsiLhat(\bz)$ differs from the full log likelihood.
As is well known the exact posterior for linear regression is given by 
$\NN(\bmu_\post, \bLam_\post^{-1})$ where $\bmu_\post = \bLam_\post^{-1} (\bLam_0 \bmu_0 + \frac{1}{\sigmaobs^2} \bX^\top \by)$ 
and $\bLam_\post = \bLam_0 + \frac{1}{\sigmaobs^2} \bX^\top \bX$.
The gradient of the full log likelihood is given by 
%%%
\begin{equation}
    \nabla_\bz \PsiL(\DD, \bz) = \frac{1}{\sigmaobs^2} \left( \by^\top \bX - \bX^\top \bX \bz \right) =  \bA - \bB \bz
\end{equation}
%%%
where we have defined $\bA \equiv \frac{1}{\sigmaobs^2}\by^\top \bX $ and $\bB \equiv \frac{1}{\sigmaobs^2}\bX^\top \bX$.
We suppose that $\PsiLhat(\bz)$ is likewise a
quadratic\footnote{Note that $\PsiLhat(\bz)$ will be precisely of this form if we use the \texttt{RAND} parameterization.
In other words Eqn.~\ref{eqn:blrsurr} follows directly from Eqn.~\ref{eqn:surrogateansatz}, which is agnostic to the particular
likelihood that appears in the model.} function of $\bz$
but that $\PsiLhat(\bz) \ne \PsiL(\DD, \bz)$  so that we can write
%%%
\begin{equation}
\label{eqn:blrsurr}
    \begin{split}
    \nabla_\bz \PsiLhat(\bz) = \bA + \dbA - (\bB + \dbB) \bz 
    \end{split}
\end{equation}
%%%
%%%
%\begin{equation}
%\begin{split}
%    \log p(\DD) - \LL_\text{DAIS} = \\ \frac{1}{2} \|\bmu_K - \bmu_\post \|_{\bLam_\post}^2  + \frac{1}{2}\trace(\bLam_\post \bSig_K) - \frac{d}{2}  + \\
%    \frac{1}{2}\log \frac{|\bSig_\post|}{|\bSig_0|} - \EE_{q} \left[ \sum_{k=1}^K  \log \frac{\pi(\hat{v}_k)}{\pi(v_{k-1})} \right]
%\end{split}
%\end{equation}
%%%
For $\nabla_\bz \PsiLhat(\bz)$ as in Eqn.~\ref{eqn:blrsurr} we can prove the following: % convergence result.
%%%
\begin{restatable}{prop}{propsldais}
\label{prop:sldais}
Running SL-DAIS for linear regression with $K$ steps under Assumption~\ref{ass} and using $\nabla_\bz \PsiLhat(\bz)$ as  
in Eqn.~\ref{eqn:blrsurr} results in a variational gap that can be bounded as
%%%
\begin{equation}
\begin{split}
\nonumber
    \log p(\DD) - \LL_\text{SL-DAIS} \le \underbrace{\OO(K^{-1/2})}_{\rm DAIS \; error} \; + \!\!\!
    \underbrace{ | \trace \bLam^{-1}_\post \dbB |}_{\rm surrogate\; likelihood \; error}
\end{split}
\end{equation}
where we have dropped higher order terms in $\dbA$ and $\dbB$.\footnote{$\dbA$ first appears at second order.
Note that we drop higher order terms simply to make the bound more readily interpretable.}
Furthermore the KL divergence between $\qf(\bz_K)$ and the posterior $\pth(\bz_K | \DD)$ is bounded
by the same quantity.
%%%
\end{restatable}
%%%
This intuitive result can be proven by suitably adapting the results in \citet{zhang2021differentiable};
see Sec.~\ref{app:sldais} in the supplemental materials for details.

Proposition \ref{prop:sldais} suggests that SL-DAIS offers an intuitive trade-off between inference speed and fidelity. For example, for a fixed computational cost,
SL-DAIS can accommodate a value of $K$ that is a factor $\sim N/B$ larger than DAIS.
For a sufficiently good surrogate log likelihood, the benefits of larger $K$ can more than compensate for the error introduced by an imperfect $\PsiLhat(\bz)$.
See Sec.~\ref{sec:logistic} for concrete examples.

%%%%%%%%%%%%%%%%%%%%%%%%%%%%%%%%%%%%%%%%%%%
\section{Related work}
\label{sec:related}
%%%%%%%%%%%%%%%%%%%%%%%%%%%%%%%%%%%%%%%%%%%

Many variational objectives that leverage importance sampling (IS) have been proposed.
These include the importance weighted autoencoder (IWAE) \citep{burda2015importance,cremer2017reinterpreting},
the thermodynamic variational objective \citep{masrani2019thermodynamic},
and approaches that make use of Sequential Monte Carlo \citep{le2017auto,maddison2017filtering,naesseth2018variational}.
For a general discussion of IS in variational inference see \citet{domke2018importance}.

An early combination of MCMC methods with variational inference was proposed by \citet{salimans2015markov} and 
\citet{wolf2016variational}. A disadvantage of these approaches is the need to learn reverse kernels, 
a shortcoming that was later addressed by \citet{caterini2018hamiltonian}.

Bayesian coresets enable users to run MCMC on large datasets after first distilling the data into a smaller 
number of weighted data points \citep{huggins2016coresets, campbell2018bayesian, campbell2019automated}.
These methods do not support model learning.
Finally a number of authors have explored MCMC methods that enable data subsampling 
\citep{maclaurin2015firefly, quiroz2018speeding, zhang2020asymptotically},
including those that leverage stochastic gradients \citep{welling2011bayesian,chen2014stochastic,ma2015complete}.
See Sec.~\ref{app:related} for an extended discussion of related work.

%%%%%%%%%%%%%%%%%%%%%%%%%%%%%%%%%%%%%%%%%%%
\section{Experiments}
\label{sec:exp}
%%%%%%%%%%%%%%%%%%%%%%%%%%%%%%%%%%%%%%%%%%%

Next we compare the performance of SL-DAIS and NL-DAIS to various MCMC and variational
baselines.
Our experiments are implemented using JAX \citep{bradbury2020jax} and NumPyro \citep{phan2019composable,bingham2019pyro}.
An open source implementation of our method will be made available at 
\scalebox{1.25}{{\scriptsize \texttt{https://num.pyro.ai/en/stable/autoguide.html}}}.

%%%%%%%%%%%%%%%%%%%%
\subsection{Surrogate log likelihood comparison}
\label{sec:ablation}
%%%%%%%%%%%%%%%%%%%%

We compare four ans\"atze for the surrogate log likelihood $\PsiLhat$ used in SL-DAIS.
For concreteness we consider a logistic regression model with a bernoulli likelihood $p(y_n | \bz, \bx_n)$
governed by logits $\sigma(\bx_n \cdot \bz)$, where $\sigma(\cdot)$ is the logistic function.
In \texttt{RAND} we randomly choose $\Nsurr$ surrogate data points $\{ (\ty_n, \tbx_n) \} \subset \DD$, introduce
a $\Nsurr$-dimensional vector of learnable weights $\bomega$ and let $\PsiLhat(\bz) = \sum_n \omega_n \log p(\ty_n | \bz, \tbx_n)$.
In \texttt{CS-INIT} we proceed similarly but use a Bayesian coreset algorithm \citep{huggins2016coresets} to
choose the $\Nsurr$ surrogate data points.
In \texttt{CS-FIX} we likewise use a coreset algorithm to choose the surrogate data points but instead of learning $\bomega$
we use the weights provided by the coreset algorithm.
Finally in $\texttt{NN}$ we parameterize $\PsiLhat(\bz)$ as a neural network.
See Table~\ref{table:ablation} for the results. 

We can read off several conclusions from Table~\ref{table:ablation}.
First, using a neural ansatz works extremely poorly. This is not surprising, since
a neural ansatz does not leverage the known likelihood function.
Second, for the three methods that rely on surrogate data points $\{ (\ty_n, \tbx_n) \} $, results
improve as we increase $\Nsurr$, although the improvements are somewhat modest for the two inference problems we consider.
Third, the simplest ansatz, namely \texttt{RAND}, works quite well.
This is encouraging because this ansatz is simple, generic, and easily automated,
which makes it an excellent candidate for probabilistic programming frameworks.
Consequently we use \texttt{RAND} in all subsequent experiments. 
Finally, the poor performance of $\texttt{CS-FIX}$ makes it clear that the notion of Bayesian coresets,
while conceptually similar, is not congruent with our use case for surrogate likelihoods. 
In particular the coreset notion in \cite{huggins2016coresets} aims to approximate $\PsiL(\DD, \bz)$ across
latent space as a whole. However, the zero-avoiding behavior of variational inference implies that reproducing 
the precise tail behavior of $\PsiL(\DD, \bz)$ is less important than accurately representing the vicinity of the posterior mode. 
Additionally the Hamiltonian `integration error' resulting from using finite $\eta$ and $K$ can be partially corrected 
for by learning $\bomega$ jointly with $q_0(\bz)$, something that a two-stage coreset-based approach is unable to do.

\newcommand{\ShiggsRAND}{{\small $211.1$}}
\newcommand{\MhiggsRAND}{{\small $214.2$}}
\newcommand{\LhiggsRAND}{{\small $222.4$}}
\newcommand{\ShiggsRANDpm}{{\small $214.3$}}
\newcommand{\MhiggsRANDpm}{{\small $217.0$}}
\newcommand{\LhiggsRANDpm}{{\small $217.8$}}
\newcommand{\ShiggsCS}{{\small $209.6$}}
\newcommand{\MhiggsCS}{{\small $213.8$}}
\newcommand{\LhiggsCS}{{\small $218.5$}}
\newcommand{\ShiggsFIXCS}{{\small $83.0$}}
\newcommand{\MhiggsFIXCS}{{\small $136.6$}}
\newcommand{\LhiggsFIXCS}{{\small $192.7$}}
\newcommand{\ShiggsCSpm}{{\small $213.5$}}
\newcommand{\MhiggsCSpm}{{\small $214.5$}}
\newcommand{\LhiggsCSpm}{{\small $219.0$}}
\newcommand{\ShiggsFIXKMpm}{{\small $218.3$}}
\newcommand{\MhiggsFIXKMpm}{{\small $217.7$}}
\newcommand{\LhiggsFIXKMpm}{{\small $217.4$}}
\newcommand{\ShiggsKMpm}{{\small $184.0$}}
\newcommand{\MhiggsKMpm}{{\small $187.4$}}
\newcommand{\LhiggsKMpm}{{\small $182.4$}}
\newcommand{\ShiggsNN}{{\small $16.8$}}
\newcommand{\MhiggsNN}{{\small $16.8$}}
\newcommand{\LhiggsNN}{{\small $16.8$}}
\newcommand{\SsusyRAND}{{\small $627.5$}}
\newcommand{\MsusyRAND}{{\small $638.9$}}
\newcommand{\LsusyRAND}{{\small $637.2$}}
\newcommand{\SsusyRANDpm}{{\small $635.0$}}
\newcommand{\MsusyRANDpm}{{\small $637.9$}}
\newcommand{\LsusyRANDpm}{{\small $632.2$}}
\newcommand{\SsusyCS}{{\small $625.9$}}
\newcommand{\MsusyCS}{{\small $645.3$}}
\newcommand{\LsusyCS}{{\small $633.7$}}
\newcommand{\SsusyFIXCS}{{\small $501.0$}}
\newcommand{\MsusyFIXCS}{{\small $538.5$}}
\newcommand{\LsusyFIXCS}{{\small $555.5$}}
\newcommand{\SsusyCSpm}{{\small $633.2$}}
\newcommand{\MsusyCSpm}{{\small $634.8$}}
\newcommand{\LsusyCSpm}{{\small $637.4$}}
\newcommand{\SsusyFIXKMpm}{{\small $635.5$}}
\newcommand{\MsusyFIXKMpm}{{\small $633.9$}}
\newcommand{\LsusyFIXKMpm}{{\small $632.1$}}
\newcommand{\SsusyKMpm}{{\small $222.3$}}
\newcommand{\MsusyKMpm}{{\small $188.3$}}
\newcommand{\LsusyKMpm}{{\small $169.7$}}
\newcommand{\SsusyNN}{{\small $2.6$}}
\newcommand{\MsusyNN}{{\small $2.6$}}
\newcommand{\LsusyNN}{{\small $2.6$}}
\newcommand{\uShiggsRAND}{{\small $211.1 \pm 1.1$}}
\newcommand{\uMhiggsRAND}{{\small $214.2 \pm 1.3$}}
\newcommand{\uLhiggsRAND}{{\small $222.4 \pm 1.1$}}
\newcommand{\uShiggsRANDpm}{{\small $214.3 \pm 1.0$}}
\newcommand{\uMhiggsRANDpm}{{\small $217.0 \pm 1.5$}}
\newcommand{\uLhiggsRANDpm}{{\small $217.8 \pm 1.5$}}
\newcommand{\uShiggsCS}{{\small $209.6 \pm 1.2$}}
\newcommand{\uMhiggsCS}{{\small $213.8 \pm 1.1$}}
\newcommand{\uLhiggsCS}{{\small $218.5 \pm 1.8$}}
\newcommand{\uShiggsFIXCS}{{\small $83.0 \pm 12.3$}}
\newcommand{\uMhiggsFIXCS}{{\small $136.6 \pm 12.6$}}
\newcommand{\uLhiggsFIXCS}{{\small $192.7 \pm 5.5$}}
\newcommand{\uShiggsCSpm}{{\small $213.5 \pm 0.9$}}
\newcommand{\uMhiggsCSpm}{{\small $214.5 \pm 1.4$}}
\newcommand{\uLhiggsCSpm}{{\small $219.0 \pm 1.7$}}
\newcommand{\uShiggsFIXKMpm}{{\small $218.3 \pm 1.3$}}
\newcommand{\uMhiggsFIXKMpm}{{\small $217.7 \pm 1.4$}}
\newcommand{\uLhiggsFIXKMpm}{{\small $217.4 \pm 1.8$}}
\newcommand{\uShiggsKMpm}{{\small $184.0 \pm 2.5$}}
\newcommand{\uMhiggsKMpm}{{\small $187.4 \pm 3.0$}}
\newcommand{\uLhiggsKMpm}{{\small $182.4 \pm 2.8$}}
\newcommand{\uShiggsNN}{{\small $16.8 \pm 0.9$}}
\newcommand{\uMhiggsNN}{{\small $16.8 \pm 0.9$}}
\newcommand{\uLhiggsNN}{{\small $16.8 \pm 0.9$}}
\newcommand{\uSsusyRAND}{{\small $627.5 \pm 2.1$}}
\newcommand{\uMsusyRAND}{{\small $638.9 \pm 9.7$}}
\newcommand{\uLsusyRAND}{{\small $637.2 \pm 2.2$}}
\newcommand{\uSsusyRANDpm}{{\small $635.0 \pm 3.0$}}
\newcommand{\uMsusyRANDpm}{{\small $637.9 \pm 4.0$}}
\newcommand{\uLsusyRANDpm}{{\small $632.2 \pm 3.8$}}
\newcommand{\uSsusyCS}{{\small $625.9 \pm 2.2$}}
\newcommand{\uMsusyCS}{{\small $645.3 \pm 2.7$}}
\newcommand{\uLsusyCS}{{\small $633.7 \pm 3.8$}}
\newcommand{\uSsusyFIXCS}{{\small $501.0 \pm 36.2$}}
\newcommand{\uMsusyFIXCS}{{\small $538.5 \pm 14.2$}}
\newcommand{\uLsusyFIXCS}{{\small $555.5 \pm 15.8$}}
\newcommand{\uSsusyCSpm}{{\small $633.2 \pm 3.7$}}
\newcommand{\uMsusyCSpm}{{\small $634.8 \pm 5.5$}}
\newcommand{\uLsusyCSpm}{{\small $637.4 \pm 3.6$}}
\newcommand{\uSsusyFIXKMpm}{{\small $635.5 \pm 3.1$}}
\newcommand{\uMsusyFIXKMpm}{{\small $633.9 \pm 2.6$}}
\newcommand{\uLsusyFIXKMpm}{{\small $632.1 \pm 2.5$}}
\newcommand{\uSsusyKMpm}{{\small $222.3 \pm 24.0$}}
\newcommand{\uMsusyKMpm}{{\small $188.3 \pm 11.4$}}
\newcommand{\uLsusyKMpm}{{\small $169.7 \pm 13.9$}}
\newcommand{\uSsusyNN}{{\small $2.6 \pm 1.4$}}
\newcommand{\uMsusyNN}{{\small $2.6 \pm 1.4$}}
\newcommand{\uLsusyNN}{{\small $2.6 \pm 1.4$}}

\begin{table}[H]
    \centering
    \resizebox{.95\columnwidth}{!}{%
\begin{tabular}{l|l|l|l|l}
\hline
{\small  Dataset} & \multicolumn{2}{l|}{{\small Higgs}} & \multicolumn{2}{l}{{\small SUSY}} \\ \hline
{\small $\Nsurr$}     & {\small $64$} & {\small $1024$} & {\small $64$} & {\small $1024$}            \\ \hline \hline
\small{\texttt{RAND}}         & \ShiggsRAND    & \LhiggsRAND    & \SsusyRAND    & \LsusyRAND    \\ \hline 
%{\small RAND$\pm$}    & \ShiggsRANDpm  & \LhiggsRANDpm  & \SsusyRANDpm  & \LsusyRANDpm  \\ \hline 
\small{\texttt{CS-INIT}}      & \ShiggsCS      & \LhiggsCS      & \SsusyCS      & \LsusyCS      \\ \hline
%{\small CS-INIT$\pm$} & \ShiggsCSpm    & \LhiggsCSpm    & \SsusyCSpm    & \LsusyCSpm    \\ \hline
\small{\texttt{CS-FIX}}       & \ShiggsFIXCS   & \LhiggsFIXCS   & \SsusyFIXCS   & \LsusyFIXCS   \\ \hline
%{\texttt{KM-INIT$\pm$} & \ShiggsKMpm    & \LhiggsKMpm    & \SsusyKMpm    & \LsusyKMpm    \\ \hline
%{\texttt{KM-FIX$\pm$}  & \ShiggsFIXKMpm & \LhiggsFIXKMpm & \SsusyFIXKMpm & \LsusyFIXKMpm \\ \hline
\small{\texttt{NN}}           & \multicolumn{2}{l|}{\ShiggsNN}  & \multicolumn{2}{l}{\SsusyNN}  \\ \hline
%{\texttt{NN}          & \ShiggsNN      & \LhiggsNN      & \SsusyNN      & \LsusyNN              
\end{tabular}
    } % end resize
\caption{We compare four ans\"atze for the surrogate log likelihood $\PsiLhat$ 
on two logistic regression datasets.
For the three strategies that make use of surrogate data points, we vary $\Nsurr \in \{64, 1024\}$.  
We report ELBO improvements (in nats) above a mean-field Normal baseline
and average results across 20 replications.
See Sec.~\ref{sec:ablation} for details and Table~\ref{table:ablation_full} in the 
supplement for expanded results, including uncertainties and additional parameterizations.
}
\label{table:ablation}
\end{table}

%%%%%%%%%%%%%%%%%%%%
\subsection{Classifying imbalanced data}
\label{sec:imbalance}
%%%%%%%%%%%%%%%%%%%%

\begin{figure}[ht]
\begin{center}
\includegraphics[width=0.75\columnwidth]{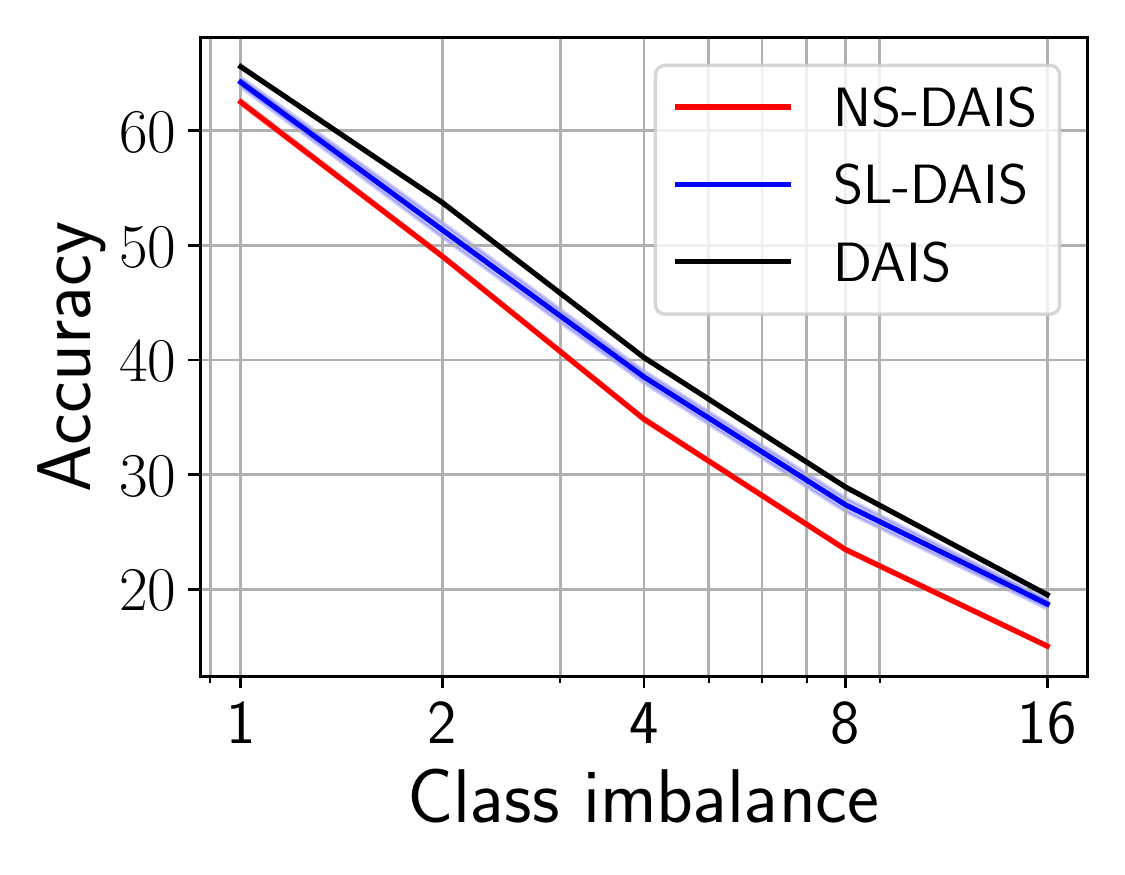}
\caption{We depict test accuracy w.r.t.~the rare class label on an imbalanced binary classification dataset.
    We compare DAIS to SL-DAIS with $\Nsurr=256$ and NS-DAIS with $B=256$.
    The horizontal axis encodes the ratio between the number of non-rare and rare class labels.
    Accuracies are averaged over 20 (3) independent replications for SL-DAIS and NS-DAIS (DAIS), respectively,
    and shaded bands denote 90\% confidence intervals.
    }
    \label{fig:imbalanced}
\end{center}
\end{figure}

To better understand the limitations of NS-DAIS and SL-DAIS we consider a binary classification
problem with imbalanced data. We expect both methods to find this regime challenging, since the
full log likelihood $\PsiL(\DD, \bz)$ exhibits elevated sensitivity to the small number of terms involving the rare class. 
See Fig.~\ref{fig:imbalanced} for the results. 
Test accuracies decrease substantially as the class imbalance increases;
this is expected, since there is considerable overlap between the two classes in feature space. Strikingly, SL-DAIS with the \texttt{RAND} surrogate ansatz
outperforms NS-DAIS across the board,\footnote{We observe similar behavior for the ELBO.}
with the gap increasing as the class imbalance increases. 
Indeed NS-DAIS barely outperforms a mean-field baseline (not shown for legibility).
This suggests that SL-DAIS $+$ \texttt{RAND} can be 
a viable strategy even in difficult regimes like the one explored here.
%%%
%\begin{wrapfigure}{L}{0.45\columnwidth}
%\captionsetup{justification=raggedright}
%\end{wrapfigure}
%%%

%%%%%%%%%%%%%%%%%%%%
\subsection{Logistic regression}
\label{sec:logistic}
%%%%%%%%%%%%%%%%%%%%

We use five logistic regression datasets to systematically compare 
eight variational methods and two MCMC methods. 
For each dataset we consider a somewhat moderate number of training data ($N\!=\!5 \times 10^4$) so that we can include baseline methods that are impractical
for larger $N$.
Our baselines include three variational methods that use parametric distributions: 
mean-field Normal (\texttt{MF});
multivariate Normal (\texttt{MVN}); and
a block neural autoregressive flow (\texttt{Flow}) \citep{de2020block}.
They also include \texttt{DAIS-MF}, \texttt{NS-DAIS-MF}, and \texttt{SL-DAIS-MF} (all of which use a mean-field base distribution) and \texttt{SL-DAIS-MVN}, which uses a MVN base distribution. NS-DAIS and SL-DAIS use $B=256$ and $\Nsurr=256$, respectively.
The two MCMC methods are \texttt{HMCECS} \citep{dang2019hamiltonian}, a variant of HMC that considers subsets of data in each iteration,
and \texttt{NUTS-CS}, where we first form a Bayesian coreset consisting of $1024$ weighted data points, and then run NUTS \citep{hoffman2014no}.
We emphasize that these two MCMC methods do not offer some of the benefits of variational methods
(e.g.~support for model learning) but we include them so that we can better assess the quality of 
the approximate posterior predictive distributions returned by the variational methods.

\begin{figure*}[ht]
\begin{center}
\includegraphics[width=1.99\columnwidth]{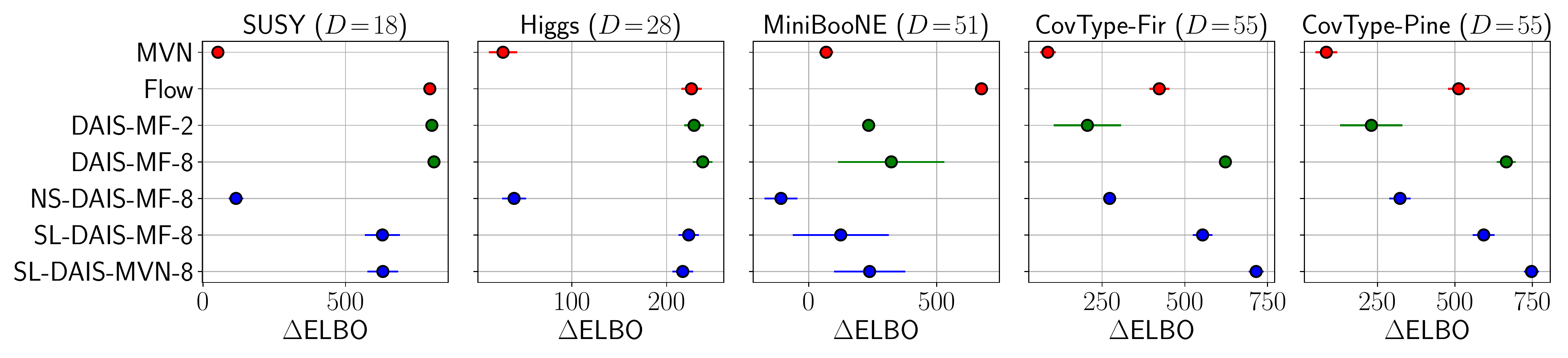}
\includegraphics[width=1.99\columnwidth]{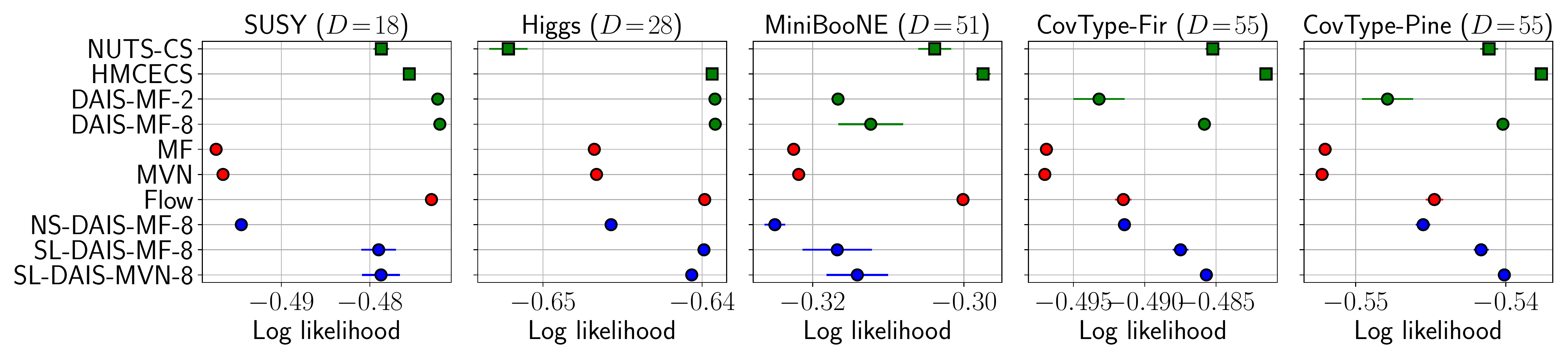}
\caption{We report ELBO improvements and  test log likelihoods for the logistic regression experiment in Sec.~\ref{sec:logistic}.
         ELBO improvements are with respect to the mean-field (MF) Normal baseline.
         Circles denote variational methods and squares denote MCMC methods. 
         Blue methods are ours, red methods are mini-batchable variational methods, and green methods are everything else.
         Metrics are averaged over 7 independent replications
         and error bars denote standard errors (we do 3 independent replications for the most expensive methods, namely DAIS and HMCECS). 
         See Sec.~\ref{app:logistic} in the supplement for test accuracies. 
         Note that numerals in method names indicate the number of HMC steps $K$ used.  }
    \label{fig:logistic}
\end{center}
\end{figure*}

\begin{figure}[ht]
\begin{center}
\includegraphics[width=0.85\columnwidth]{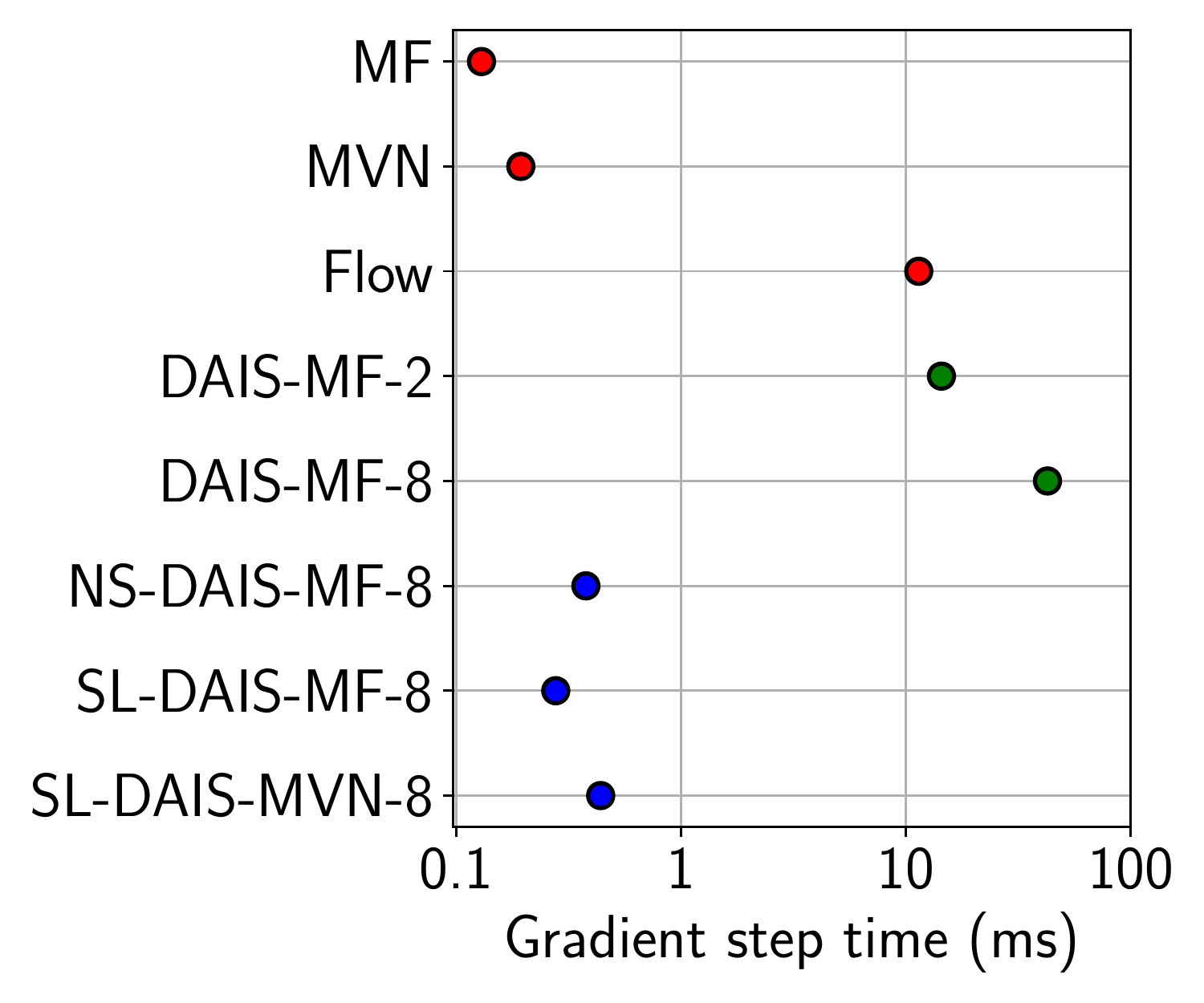}
    \caption{We report the time per optimization step for each variational method in Fig.~\ref{fig:logistic}
    on the CovType-Fir dataset. Runtimes are for a CPU with 24 cores (Intel Xeon Gold 5220R 2.2GHz). 
    }
    \label{fig:timingcpu}
\end{center}
\end{figure}

See Fig.~\ref{fig:logistic} and Table~\ref{table:rank} for results.
We find that SL-DAIS consistently outperforms NS-DAIS and in many cases it
matches the performance of DAIS. Among the scalable methods \texttt{Flow} and SL-DAIS perform best,
although SL-DAIS can be substantially faster than \texttt{Flow}, see Fig.~\ref{fig:timingcpu}. 
Fig.~\ref{fig:timingcpu} also clarifies the computational benefits of SL-DAIS: 
SL-DAIS with $K=8$ steps handily outperforms DAIS with $K=2$ at about $2\%$ of the computational cost.
Finally we find that the best variational methods yield predictive log likelihoods that are competitive
with the MCMC baselines.
%See Table~\ref{table:rank} for performance ranks for all the scalable variational methods we consider.

%%%
%%%

\begin{table*}[t]
    \centering
    \resizebox{0.99\linewidth}{!}{%
\begin{tabular}{llllllll}
\hline
\small{}         & \small{MF} & \small{MVN} & \small{Flow} & \small{NS-DAIS-MF} & \small{SL-DAIS-MF} & \small{NS-DAIS-MVN} & \small{SL-DAIS-MVN}  \\ \hline
    \small{ELBO}           & \small{$6.60$} & \small{$5.36$} & \small{\bf 2.08} & \small{$5.00$} & \small{$2.60$} & \small{$4.16$} & \small{$2.20$}  \\
\small{Log likelihood} & \small{$6.16$} & \small{$5.88$} & \small{\bf 2.16} & \small{$4.60$} & \small{$2.44$} & \small{$4.36$} & \small{$2.40$} \\
%\small{Accuracy}       & \small{$5.48$} & \small{$5.24$} & \small{$1.96$} & \small{$4.52$} & \small{$3.40$} & \small{$4.12$} & \small{$3.28$} \\ \hline
\small{Opt. time}       & \small{\bf 38.8} & \small{$58.1$} & \small{$3420.3$} & \small{$113.1$} & \small{$83.1$} & \small{$152.1$} & \small{$131.6$} \\ \hline
\end{tabular}
     } % end resize
\caption{We report performance ranks w.r.t.~ELBO and test log likelihood across 5 train/test splits and 5 datasets for the logistic regression
    experiment in Sec.~\ref{sec:logistic}. We also report time per optimization step in milliseconds as in Fig.~\ref{fig:timingcpu}.
    Lower is better for all metrics. The rank satisfies $1 \le {\rm rank} \le 7$, since we compare 7 scalable variational methods.}
\label{table:rank}
\end{table*}

%%%
%%%

%%%%%%%%%%%%%%%%%%%%
\subsection{Robust Gaussian process regression}
\label{sec:robust}
%%%%%%%%%%%%%%%%%%%%

We model a geospatial precipitation dataset \cite{lyon2004strength,lyon2005enso} considered in \cite{pleiss2020fast}
using a Gaussian process regressor with a heavy-tailed Student's t likelihood; see Sec.~\ref{app:robust} for details
on the modeling setup. 
We compare SL-DAIS with $K=4$, $\Nsurr=256$, and a multivariate Normal (MVN) base distribution $q_0(\bz)$ to a baseline with a MVN variational distribution.
Note that the dimension of the latent space $D$ is equal to the number of inducing points.
See Table~\ref{table:robust} for the results.

There is significant variability in the data and consequently the posterior function values exhibit considerable uncertainty.
Due to the non-conjugate likelihood, the posterior over the inducing points features significant non-gaussianity,
especially as the number of inducing points increases and the inducing point locations become closer together. 
Consequently we see the largest ELBO and test log likelihood improvements for the largest number of inducing points.  

\begin{table}[H]
    \centering
    \resizebox{\columnwidth}{!}{%
\begin{tabular}{l|l|l|l}
\hline
%\small{\# of ind. points} & \small{64}                 & \small{128}              & \small{256} \\ \hline 
\small{latent dim.}  & \small{$D=64$}             & \small{$D=128$}           & \small{$D=256$}           \\ \hline 
\small{$\Delta$ELBO} & \small{$127.3 \pm 26.6$}   & \small{$187.1 \pm 26.6$}  & \small{$529.5 \pm 17.3$}  \\ \hline
\small{$\Delta$LL}   & \small{$0.014 \pm 0.003$}  & \small{$0.015 \pm 0.003$} & \small{$0.039 \pm 0.006$} \\ \hline
\end{tabular}
     } % end resize
\caption{We report ELBO and test log likelihood (LL) improvements in nats for the regression experiment
in Sec.~\ref{sec:robust}, comparing SL-DAIS to a baseline with a multivariate Normal variational distribution. 
Test log likelihoods are normalized by the number of test points. Results are averaged over seven train/test splits.} 
\label{table:robust}
\end{table}
%	elbo[ni=64]	 \small{$127.3 \pm 26.6$}
%	ll[ni=64]  	 \small{$0.014 \pm 0.003$}
%	elbo[ni=128] \small{$187.1 \pm 26.6$}
%	ll[ni=128]   \small{$0.015 \pm 0.003$}
%	elbo[ni=256] \small{$529.5 \pm 17.3$}
%	ll[ni=256]   \small{$0.039 \pm 0.006$}

%%%%%%%%%%%%%%%%%%%%
\subsection{Gaussian process classification}
\label{sec:fitc}
%%%%%%%%%%%%%%%%%%%%

We train Gaussian process classifiers on the classification datasets considered in Sec.~\ref{sec:logistic}.
We compare SL-DAIS with $K=4$, $\Nsurr=512$, and a MVN base distribution to a baseline with a MVN variational distribution.
We use $128$ inducing points so the latent dimension is $D=128$.
See Table~\ref{table:fitc} for results. We find that SL-DAIS leads to consistently higher ELBOs
and that this translates to small but non-negligible gains in test accuracy for the datasets
with the largest ELBO gains. 

We emphasize that this and the preceding Gaussian process experiments represent
difficult inference problems, since the latent dimensionality is moderately high (with $D$ as large as $256$)
and since high-dimensional model parameters $\theta$ (including $D$ inducing point locations) are learned jointly
with the approximate posterior.

\begin{table}[H]
    \centering
    \resizebox{.95\columnwidth}{!}{%
\begin{tabular}{l|l|l}
\hline
\small{Dataset}      & \small{$\Delta$ELBO}     & \small{$\Delta$Accuracy}  \\ \hline 
\small{SUSY}         & \small{$34.0 \pm 2.2$}   & \small{$0.002 \pm 0.019$} \\ \hline
\small{Higgs}        & \small{$225.5 \pm 7.3$}  & \small{$0.072 \pm 0.044$} \\ \hline
\small{MiniBooNE}    & \small{$163.9 \pm 5.4$}  & \small{$0.121 \pm 0.011$} \\ \hline
\small{CovType-Fir}  & \small{$471.7 \pm 42.2$} & \small{$0.399 \pm 0.080$} \\ \hline
\small{CovType-Pine} & \small{$616.6 \pm 71.8$} & \small{$0.425 \pm 0.045$} \\ \hline
\end{tabular}
     } % end resize
\caption{We report ELBO and test accuracy improvements for the classification experiment
in Sec.~\ref{sec:fitc}, comparing SL-DAIS to a baseline with a multivariate Normal variational distribution. 
Results are averaged over five train/test splits.} 
\label{table:fitc}
\end{table}

%  elbo[susy]         	 \small{$34.0 \pm 2.2$}
%  ll[susy]         	 \small{$0.0004 \pm 0.00005$}
%  acc[susy]         	 \small{$0.002 \pm 0.019$}

%  elbo[higgs]        	 \small{$225.5 \pm 7.3$}
%  ll[higgs]        	 \small{$0.0003 \pm 0.00023$}
%  acc[higgs]        	 \small{$0.072 \pm 0.044$}

%  elbo[miniboone]    	 \small{$163.9 \pm 5.4$}
%  ll[miniboone]    	 \small{$0.0024 \pm 0.00020$}
%  acc[miniboone]    	 \small{$0.121 \pm 0.011$}

%  elbo[covtype-fir]  	 \small{$471.7 \pm 42.2$}
%  ll[covtype-fir]  	 \small{$0.0085 \pm 0.00103$}
%  acc[covtype-fir]  	 \small{$0.399 \pm 0.080$}

%  elbo[covtype-pine] 	 \small{$616.6 \pm 71.8$}
%  ll[covtype-pine] 	 \small{$0.0112 \pm 0.00144$}
%  acc[covtype-pine] 	 \small{$0.425 \pm 0.045$}

%%%%%%%%%%%%%%%%%%%%
\subsection{Local latent variable models}
\label{sec:llv}
%%%%%%%%%%%%%%%%%%%%

What about models with global and local latent variables?
For simplicity we evaluate the performance of the simplest DAIS-like inference procedure
that can simultaneously accommodate local latent variables and data subsampling.
In brief we introduce a parametric distribution for the global latent variable and use
DAIS to define a distribution over the local latent variables. Assuming the local latent
variables are conditionally independent once we condition on the global latent variable, 
this results in $N$ \emph{non-interacting} DAIS chains.
Consequently the ELBO is amenable to data subsampling; see Sec.~\ref{app:llvm} for an extended discussion

To evaluate this approach we consider a robust linear regression model that uses a Student's t likelihood.
We use the well-known representation of this likelihood as a continuous mixture of Normal distributions.
This yields a model with local Gamma variates where the local latent variables can be integrated out exactly.

We compare two variational approaches, both of which use
a mean-field Normal distribution for the global latent variable. 
To define an oracle baseline we integrate out the local latent variables before performing variational inference. 
This oracle represents an upper performance bound on the semi-parametric approach defined above, which we refer to as Semi-DAIS.
See Table~\ref{table:local} for results. We find that for $K=16$ Semi-DAIS `recovers' about half of the gap between
a fully mean-field baseline and the oracle. We expect this gap would decrease further for larger $K$.

\begin{table}[H]
    \centering
    \resizebox{.99\columnwidth}{!}{%
\begin{tabular}{l|l|l|l}
\hline
\small{Dataset}   & \small{Semi-DAIS-8}     & \small{Semi-DAIS-16}    & \small{Oracle}          \\ \hline 
\small{Pol}       & \small{$91.4 \pm 3.3$}  & \small{$136.4 \pm 3.6$} & \small{$360.2 \pm 2.6$} \\ \hline
\small{Elevators} & \small{$186.3 \pm 5.3$} & \small{$366.8 \pm 5.5$} & \small{$758.0 \pm 2.5$} \\ \hline
\end{tabular}
     } % end resize
    \caption{We report results for the experiment in Sec.~\ref{sec:llv}.
    In each case the reported ELBO improvement is above a mean field baseline.
    Results are averaged over 10 replications.}
\label{table:local}
\end{table}

%%%%% pol
%	elbo[SEMIDAIS-8]  	 \small{$91.4 \pm 3.3$}
%	elbo[SEMIDAIS-16]  	 \small{$136.4 \pm 3.6$}
%	elbo[ORACLE]      	 \small{$360.2 \pm 2.6$}

%%%%% elevators
%	elbo[SEMIDAIS-8]  	 \small{$186.3 \pm 5.3$}
%	elbo[SEMIDAIS-16]  	 \small{$366.8 \pm 5.5$}
%	elbo[ORACLE]      	 \small{$758.0 \pm 2.5$}

%%%%%%%%%%%%%%%%%%%%%%%%%%%%%%%%%%%%%%%%%%%
\section{Discussion}
\label{sec:disc}
%%%%%%%%%%%%%%%%%%%%%%%%%%%%%%%%%%%%%%%%%%%

In this work we have focused on models with global latent variables.
The experiment in Sec.~\ref{sec:llv} only scratches the surface of what is possible
for the richer and more complex case of models that contain global and local latent variables.
Exploring hybrid scalable inference strategies for this class of models that combine gradient-based MCMC with variational methods 
is an interesting direction for future work.

%%%%%%%%%%%%%%%%%%%%%%%%%%%%%%%%%%%%%%%%%%%
\section*{Acknowledgements}

We thank Tomas Geffner for answering questions about \citet{geffner2021mcmc}.
We thank Ola Rønning for contributing the open source NumPyro-based HMCECS implementation we used in our experiments.

%%%%%%%%%%%%%%%%%%%%%%%%%%%%%%%%%%%%%%%%%%%
\bibliography{ref}

\begin{thebibliography}{62}
\providecommand{\natexlab}[1]{#1}
\providecommand{\url}[1]{\texttt{#1}}
\expandafter\ifx\csname urlstyle\endcsname\relax
  \providecommand{\doi}[1]{doi: #1}\else
  \providecommand{\doi}{doi: \begingroup \urlstyle{rm}\Url}\fi

\bibitem[Asuncion \& Newman(2007)Asuncion and Newman]{asuncion2007uci}
Asuncion, A. and Newman, D.
\newblock Uci machine learning repository, 2007.

\bibitem[Bachem et~al.(2017)Bachem, Lucic, and Krause]{bachem2017practical}
Bachem, O., Lucic, M., and Krause, A.
\newblock Practical coreset constructions for machine learning.
\newblock \emph{arXiv preprint arXiv:1703.06476}, 2017.

\bibitem[Baldi et~al.(2014)Baldi, Sadowski, and Whiteson]{baldi2014searching}
Baldi, P., Sadowski, P., and Whiteson, D.
\newblock Searching for exotic particles in high-energy physics with deep
  learning.
\newblock \emph{Nature communications}, 5\penalty0 (1):\penalty0 1--9, 2014.

\bibitem[Betancourt(2015)]{betancourt2015fundamental}
Betancourt, M.
\newblock The fundamental incompatibility of scalable hamiltonian monte carlo
  and naive data subsampling.
\newblock In \emph{International Conference on Machine Learning}, pp.\
  533--540. PMLR, 2015.

\bibitem[Betancourt(2017)]{betancourt2017conceptual}
Betancourt, M.
\newblock A conceptual introduction to hamiltonian monte carlo.
\newblock \emph{arXiv preprint arXiv:1701.02434}, 2017.

\bibitem[Bingham et~al.(2019)Bingham, Chen, Jankowiak, Obermeyer, Pradhan,
  Karaletsos, Singh, Szerlip, Horsfall, and Goodman]{bingham2019pyro}
Bingham, E., Chen, J.~P., Jankowiak, M., Obermeyer, F., Pradhan, N.,
  Karaletsos, T., Singh, R., Szerlip, P., Horsfall, P., and Goodman, N.~D.
\newblock Pyro: Deep universal probabilistic programming.
\newblock \emph{The Journal of Machine Learning Research}, 20\penalty0
  (1):\penalty0 973--978, 2019.

\bibitem[Blackard \& Dean(1999)Blackard and Dean]{blackard1999comparative}
Blackard, J.~A. and Dean, D.~J.
\newblock Comparative accuracies of artificial neural networks and discriminant
  analysis in predicting forest cover types from cartographic variables.
\newblock \emph{Computers and electronics in agriculture}, 24\penalty0
  (3):\penalty0 131--151, 1999.

\bibitem[Blei et~al.(2017)Blei, Kucukelbir, and McAuliffe]{blei2017variational}
Blei, D.~M., Kucukelbir, A., and McAuliffe, J.~D.
\newblock Variational inference: A review for statisticians.
\newblock \emph{Journal of the American statistical Association}, 112\penalty0
  (518):\penalty0 859--877, 2017.

\bibitem[Bradbury et~al.(2020)Bradbury, Frostig, Hawkins, Johnson, Leary,
  Maclaurin, and Wanderman-Milne]{bradbury2020jax}
Bradbury, J., Frostig, R., Hawkins, P., Johnson, M.~J., Leary, C., Maclaurin,
  D., and Wanderman-Milne, S.
\newblock Jax: composable transformations of python+ numpy programs, 2018.
\newblock \emph{URL http://github. com/google/jax}, 4:\penalty0 16, 2020.

\bibitem[Burda et~al.(2015)Burda, Grosse, and
  Salakhutdinov]{burda2015importance}
Burda, Y., Grosse, R., and Salakhutdinov, R.
\newblock Importance weighted autoencoders.
\newblock \emph{arXiv preprint arXiv:1509.00519}, 2015.

\bibitem[Campbell \& Broderick(2018)Campbell and
  Broderick]{campbell2018bayesian}
Campbell, T. and Broderick, T.
\newblock Bayesian coreset construction via greedy iterative geodesic ascent.
\newblock In \emph{International Conference on Machine Learning}, pp.\
  698--706. PMLR, 2018.

\bibitem[Campbell \& Broderick(2019)Campbell and
  Broderick]{campbell2019automated}
Campbell, T. and Broderick, T.
\newblock Automated scalable bayesian inference via hilbert coresets.
\newblock \emph{The Journal of Machine Learning Research}, 20\penalty0
  (1):\penalty0 551--588, 2019.

\bibitem[Caterini et~al.(2018)Caterini, Doucet, and
  Sejdinovic]{caterini2018hamiltonian}
Caterini, A.~L., Doucet, A., and Sejdinovic, D.
\newblock Hamiltonian variational auto-encoder.
\newblock \emph{arXiv preprint arXiv:1805.11328}, 2018.

\bibitem[Chen et~al.(2022)Chen, Xu, and Campbell]{chen2022bayesian}
Chen, N., Xu, Z., and Campbell, T.
\newblock Bayesian inference via sparse hamiltonian flows.
\newblock \emph{arXiv preprint arXiv:2203.05723}, 2022.

\bibitem[Chen et~al.(2014)Chen, Fox, and Guestrin]{chen2014stochastic}
Chen, T., Fox, E., and Guestrin, C.
\newblock Stochastic gradient hamiltonian monte carlo.
\newblock In \emph{International conference on machine learning}, pp.\
  1683--1691. PMLR, 2014.

\bibitem[Cremer et~al.(2017)Cremer, Morris, and
  Duvenaud]{cremer2017reinterpreting}
Cremer, C., Morris, Q., and Duvenaud, D.
\newblock Reinterpreting importance-weighted autoencoders.
\newblock \emph{arXiv preprint arXiv:1704.02916}, 2017.

\bibitem[Dang et~al.(2019)Dang, Quiroz, Kohn, Minh-Ngoc, and
  Villani]{dang2019hamiltonian}
Dang, K.-D., Quiroz, M., Kohn, R., Minh-Ngoc, T., and Villani, M.
\newblock Hamiltonian monte carlo with energy conserving subsampling.
\newblock \emph{Journal of machine learning research}, 20, 2019.

\bibitem[Dayan et~al.(1995)Dayan, Hinton, Neal, and Zemel]{dayan1995helmholtz}
Dayan, P., Hinton, G.~E., Neal, R.~M., and Zemel, R.~S.
\newblock The helmholtz machine.
\newblock \emph{Neural computation}, 7\penalty0 (5):\penalty0 889--904, 1995.

\bibitem[De~Cao et~al.(2020)De~Cao, Aziz, and Titov]{de2020block}
De~Cao, N., Aziz, W., and Titov, I.
\newblock Block neural autoregressive flow.
\newblock In \emph{Uncertainty in Artificial Intelligence}, pp.\  1263--1273.
  PMLR, 2020.

\bibitem[Ding \& Freedman(2019)Ding and Freedman]{ding2019learning}
Ding, X. and Freedman, D.~J.
\newblock Learning deep generative models with annealed importance sampling.
\newblock \emph{arXiv preprint arXiv:1906.04904}, 2019.

\bibitem[Domke \& Sheldon(2018)Domke and Sheldon]{domke2018importance}
Domke, J. and Sheldon, D.
\newblock Importance weighting and variational inference.
\newblock \emph{arXiv preprint arXiv:1808.09034}, 2018.

\bibitem[Duane et~al.(1987)Duane, Kennedy, Pendleton, and
  Roweth]{duane1987hybrid}
Duane, S., Kennedy, A.~D., Pendleton, B.~J., and Roweth, D.
\newblock Hybrid monte carlo.
\newblock \emph{Physics letters B}, 195\penalty0 (2):\penalty0 216--222, 1987.

\bibitem[Geffner \& Domke(2021)Geffner and Domke]{geffner2021mcmc}
Geffner, T. and Domke, J.
\newblock Mcmc variational inference via uncorrected hamiltonian annealing.
\newblock \emph{Advances in Neural Information Processing Systems}, 34, 2021.

\bibitem[Grosse et~al.(2015)Grosse, Ghahramani, and
  Adams]{grosse2015sandwiching}
Grosse, R.~B., Ghahramani, Z., and Adams, R.~P.
\newblock Sandwiching the marginal likelihood using bidirectional monte carlo.
\newblock \emph{arXiv preprint arXiv:1511.02543}, 2015.

\bibitem[Hoffman(2017)]{hoffman2017learning}
Hoffman, M.~D.
\newblock Learning deep latent gaussian models with markov chain monte carlo.
\newblock In \emph{International conference on machine learning}, pp.\
  1510--1519. PMLR, 2017.

\bibitem[Hoffman et~al.(2013)Hoffman, Blei, Wang, and
  Paisley]{hoffman2013stochastic}
Hoffman, M.~D., Blei, D.~M., Wang, C., and Paisley, J.
\newblock Stochastic variational inference.
\newblock \emph{Journal of Machine Learning Research}, 14\penalty0 (5), 2013.

\bibitem[Hoffman et~al.(2014)Hoffman, Gelman, et~al.]{hoffman2014no}
Hoffman, M.~D., Gelman, A., et~al.
\newblock The no-u-turn sampler: adaptively setting path lengths in hamiltonian
  monte carlo.
\newblock \emph{J. Mach. Learn. Res.}, 15\penalty0 (1):\penalty0 1593--1623,
  2014.

\bibitem[Horowitz(1991)]{horowitz1991generalized}
Horowitz, A.~M.
\newblock A generalized guided monte carlo algorithm.
\newblock \emph{Physics Letters B}, 268\penalty0 (2):\penalty0 247--252, 1991.

\bibitem[Huggins et~al.(2016)Huggins, Campbell, and
  Broderick]{huggins2016coresets}
Huggins, J., Campbell, T., and Broderick, T.
\newblock Coresets for scalable bayesian logistic regression.
\newblock \emph{Advances in Neural Information Processing Systems},
  29:\penalty0 4080--4088, 2016.

\bibitem[Jordan et~al.(1999)Jordan, Ghahramani, Jaakkola, and
  Saul]{jordan1999introduction}
Jordan, M.~I., Ghahramani, Z., Jaakkola, T.~S., and Saul, L.~K.
\newblock An introduction to variational methods for graphical models.
\newblock \emph{Machine learning}, 37\penalty0 (2):\penalty0 183--233, 1999.

\bibitem[Kingma \& Ba(2014)Kingma and Ba]{kingma2014adam}
Kingma, D.~P. and Ba, J.
\newblock Adam: A method for stochastic optimization.
\newblock \emph{arXiv preprint arXiv:1412.6980}, 2014.

\bibitem[Le et~al.(2017)Le, Igl, Rainforth, Jin, and Wood]{le2017auto}
Le, T.~A., Igl, M., Rainforth, T., Jin, T., and Wood, F.
\newblock Auto-encoding sequential monte carlo.
\newblock \emph{arXiv preprint arXiv:1705.10306}, 2017.

\bibitem[Li et~al.(2017)Li, Turner, and Liu]{li2017approximate}
Li, Y., Turner, R.~E., and Liu, Q.
\newblock Approximate inference with amortised mcmc.
\newblock \emph{arXiv preprint arXiv:1702.08343}, 2017.

\bibitem[Lyon(2004)]{lyon2004strength}
Lyon, B.
\newblock The strength of el ni{\~n}o and the spatial extent of tropical
  drought.
\newblock \emph{Geophysical Research Letters}, 31\penalty0 (21), 2004.

\bibitem[Lyon \& Barnston(2005)Lyon and Barnston]{lyon2005enso}
Lyon, B. and Barnston, A.~G.
\newblock Enso and the spatial extent of interannual precipitation extremes in
  tropical land areas.
\newblock \emph{Journal of Climate}, 18\penalty0 (23):\penalty0 5095--5109,
  2005.

\bibitem[Ma et~al.(2015)Ma, Chen, and Fox]{ma2015complete}
Ma, Y.-A., Chen, T., and Fox, E.~B.
\newblock A complete recipe for stochastic gradient mcmc.
\newblock \emph{arXiv preprint arXiv:1506.04696}, 2015.

\bibitem[Maclaurin \& Adams(2015)Maclaurin and Adams]{maclaurin2015firefly}
Maclaurin, D. and Adams, R.~P.
\newblock Firefly monte carlo: Exact mcmc with subsets of data.
\newblock In \emph{Twenty-Fourth International Joint Conference on Artificial
  Intelligence}, 2015.

\bibitem[Maddison et~al.(2017)Maddison, Lawson, Tucker, Heess, Norouzi, Mnih,
  Doucet, and Teh]{maddison2017filtering}
Maddison, C.~J., Lawson, D., Tucker, G., Heess, N., Norouzi, M., Mnih, A.,
  Doucet, A., and Teh, Y.~W.
\newblock Filtering variational objectives.
\newblock \emph{arXiv preprint arXiv:1705.09279}, 2017.

\bibitem[Masrani et~al.(2019)Masrani, Le, and Wood]{masrani2019thermodynamic}
Masrani, V., Le, T.~A., and Wood, F.
\newblock The thermodynamic variational objective.
\newblock \emph{arXiv preprint arXiv:1907.00031}, 2019.

\bibitem[Miller et~al.(2020)Miller, Foti, Lewnard, Jewell, Guestrin, and
  Fox]{miller2020mobility}
Miller, A.~C., Foti, N.~J., Lewnard, J.~A., Jewell, N.~P., Guestrin, C., and
  Fox, E.~B.
\newblock Mobility trends provide a leading indicator of changes in sars-cov-2
  transmission.
\newblock \emph{MedRxiv}, 2020.

\bibitem[Monod et~al.(2021)Monod, Blenkinsop, Xi, Hebert, Bershan, Tietze,
  Baguelin, Bradley, Chen, Coupland, et~al.]{monod2021age}
Monod, M., Blenkinsop, A., Xi, X., Hebert, D., Bershan, S., Tietze, S.,
  Baguelin, M., Bradley, V.~C., Chen, Y., Coupland, H., et~al.
\newblock Age groups that sustain resurging covid-19 epidemics in the united
  states.
\newblock \emph{Science}, 371\penalty0 (6536):\penalty0 eabe8372, 2021.

\bibitem[Naesseth et~al.(2018)Naesseth, Linderman, Ranganath, and
  Blei]{naesseth2018variational}
Naesseth, C., Linderman, S., Ranganath, R., and Blei, D.
\newblock Variational sequential monte carlo.
\newblock In \emph{International conference on artificial intelligence and
  statistics}, pp.\  968--977. PMLR, 2018.

\bibitem[Neal(2001)]{neal2001annealed}
Neal, R.~M.
\newblock Annealed importance sampling.
\newblock \emph{Statistics and computing}, 11\penalty0 (2):\penalty0 125--139,
  2001.

\bibitem[Neal et~al.(2011)]{neal2011mcmc}
Neal, R.~M. et~al.
\newblock Mcmc using hamiltonian dynamics.
\newblock \emph{Handbook of markov chain monte carlo}, 2\penalty0
  (11):\penalty0 2, 2011.

\bibitem[Obermeyer et~al.(2022)Obermeyer, Jankowiak, Barkas, Schaffner, Pyle,
  Yurkovetskiy, Bosso, Park, Babadi, MacInnis, Luban, Sabeti, and
  Lemieux]{obermeyer2021analysis}
Obermeyer, F., Jankowiak, M., Barkas, N., Schaffner, S.~F., Pyle, J.~D.,
  Yurkovetskiy, L., Bosso, M., Park, D.~J., Babadi, M., MacInnis, B.~L., Luban,
  J., Sabeti, P.~C., and Lemieux, J.~E.
\newblock Analysis of 6.4 million sars-cov-2 genomes identifies mutations
  associated with fitness.
\newblock \emph{Science}, 2022.
\newblock \doi{10.1126/science.abm1208}.
\newblock URL \url{https://www.science.org/doi/abs/10.1126/science.abm1208}.

\bibitem[Phan et~al.(2019)Phan, Pradhan, and Jankowiak]{phan2019composable}
Phan, D., Pradhan, N., and Jankowiak, M.
\newblock Composable effects for flexible and accelerated probabilistic
  programming in numpyro.
\newblock \emph{arXiv preprint arXiv:1912.11554}, 2019.

\bibitem[Pleiss et~al.(2020)Pleiss, Jankowiak, Eriksson, Damle, and
  Gardner]{pleiss2020fast}
Pleiss, G., Jankowiak, M., Eriksson, D., Damle, A., and Gardner, J.
\newblock Fast matrix square roots with applications to gaussian processes and
  bayesian optimization.
\newblock \emph{Advances in Neural Information Processing Systems},
  33:\penalty0 22268--22281, 2020.

\bibitem[Quinonero-Candela \& Rasmussen(2005)Quinonero-Candela and
  Rasmussen]{quinonero2005unifying}
Quinonero-Candela, J. and Rasmussen, C.~E.
\newblock A unifying view of sparse approximate gaussian process regression.
\newblock \emph{The Journal of Machine Learning Research}, 6:\penalty0
  1939--1959, 2005.

\bibitem[Quiroz et~al.(2018)Quiroz, Kohn, Villani, and
  Tran]{quiroz2018speeding}
Quiroz, M., Kohn, R., Villani, M., and Tran, M.-N.
\newblock Speeding up mcmc by efficient data subsampling.
\newblock \emph{Journal of the American Statistical Association}, 2018.

\bibitem[Ranganath et~al.(2014)Ranganath, Gerrish, and
  Blei]{ranganath2014black}
Ranganath, R., Gerrish, S., and Blei, D.
\newblock Black box variational inference.
\newblock In \emph{Artificial intelligence and statistics}, pp.\  814--822.
  PMLR, 2014.

\bibitem[Rezende \& Mohamed(2015)Rezende and Mohamed]{rezende2015variational}
Rezende, D. and Mohamed, S.
\newblock Variational inference with normalizing flows.
\newblock In \emph{International conference on machine learning}, pp.\
  1530--1538. PMLR, 2015.

\bibitem[Roe et~al.(2005)Roe, Yang, Zhu, Liu, Stancu, and
  McGregor]{roe2005boosted}
Roe, B.~P., Yang, H.-J., Zhu, J., Liu, Y., Stancu, I., and McGregor, G.
\newblock Boosted decision trees as an alternative to artificial neural
  networks for particle identification.
\newblock \emph{Nuclear Instruments and Methods in Physics Research Section A:
  Accelerators, Spectrometers, Detectors and Associated Equipment},
  543\penalty0 (2-3):\penalty0 577--584, 2005.

\bibitem[Ruiz \& Titsias(2019)Ruiz and Titsias]{ruiz2019contrastive}
Ruiz, F. and Titsias, M.
\newblock A contrastive divergence for combining variational inference and
  mcmc.
\newblock In \emph{International Conference on Machine Learning}, pp.\
  5537--5545. PMLR, 2019.

\bibitem[Salimans et~al.(2015)Salimans, Kingma, and
  Welling]{salimans2015markov}
Salimans, T., Kingma, D., and Welling, M.
\newblock Markov chain monte carlo and variational inference: Bridging the gap.
\newblock In \emph{International Conference on Machine Learning}, pp.\
  1218--1226. PMLR, 2015.

\bibitem[Snelson \& Ghahramani(2006)Snelson and Ghahramani]{snelson2006sparse}
Snelson, E. and Ghahramani, Z.
\newblock Sparse gaussian processes using pseudo-inputs.
\newblock \emph{Advances in neural information processing systems},
  18:\penalty0 1257, 2006.

\bibitem[Sohl-Dickstein \& Culpepper(2012)Sohl-Dickstein and
  Culpepper]{sohl2012hamiltonian}
Sohl-Dickstein, J. and Culpepper, B.~J.
\newblock Hamiltonian annealed importance sampling for partition function
  estimation.
\newblock \emph{arXiv preprint arXiv:1205.1925}, 2012.

\bibitem[Thin et~al.(2021)Thin, Kotelevskii, Doucet, Durmus, Moulines, and
  Panov]{thin2021monte}
Thin, A., Kotelevskii, N., Doucet, A., Durmus, A., Moulines, E., and Panov, M.
\newblock Monte carlo variational auto-encoders.
\newblock In \emph{International Conference on Machine Learning}, pp.\
  10247--10257. PMLR, 2021.

\bibitem[Tran et~al.(2016)Tran, Kohn, Quiroz, and Villani]{tran2016block}
Tran, M.-N., Kohn, R., Quiroz, M., and Villani, M.
\newblock The block pseudo-marginal sampler.
\newblock \emph{arXiv preprint arXiv:1603.02485}, 2016.

\bibitem[Welling \& Teh(2011)Welling and Teh]{welling2011bayesian}
Welling, M. and Teh, Y.~W.
\newblock Bayesian learning via stochastic gradient langevin dynamics.
\newblock In \emph{Proceedings of the 28th international conference on machine
  learning (ICML-11)}, pp.\  681--688. Citeseer, 2011.

\bibitem[Wolf et~al.(2016)Wolf, Karl, and van~der Smagt]{wolf2016variational}
Wolf, C., Karl, M., and van~der Smagt, P.
\newblock Variational inference with hamiltonian monte carlo.
\newblock \emph{arXiv preprint arXiv:1609.08203}, 2016.

\bibitem[Zhang et~al.(2021)Zhang, Hsu, Li, Finn, and
  Grosse]{zhang2021differentiable}
Zhang, G., Hsu, K., Li, J., Finn, C., and Grosse, R.~B.
\newblock Differentiable annealed importance sampling and the perils of
  gradient noise.
\newblock \emph{Advances in Neural Information Processing Systems}, 34, 2021.

\bibitem[Zhang et~al.(2020)Zhang, Cooper, and De~Sa]{zhang2020asymptotically}
Zhang, R., Cooper, A.~F., and De~Sa, C.~M.
\newblock Asymptotically optimal exact minibatch metropolis-hastings.
\newblock \emph{Advances in Neural Information Processing Systems},
  33:\penalty0 19500--19510, 2020.

\end{thebibliography}
\bibliographystyle{icml2022}

\clearpage

\onecolumn

%%%%%%%%%%%%%%%%%%%%%%%%%%%%%%%%%%%%%%%%%%%
\appendix
%%%%%%%%%%%%%%%%%%%%%%
\section{Appendix}
\label{sec:app}
%%%%%%%%%%%%%%%%%%%%%%

The appendix is organized as follows.
In Sec.~\ref{app:llvm} we discuss local latent variable models.
In Sec.~\ref{app:related} we provide an extended discussion of related work. 
In Sec.~\ref{app:daiselbo} we provide a proof of Lemma \ref{lemma:one}. 
In Sec.~\ref{app:nssldais} we discuss general properties of NS-DAIS and SL-DAIS. 
In Sec.~\ref{app:nsdais} we discuss NS-DAIS and 
in Sec.~\ref{app:sldais} we discuss SL-DAIS.
In Sec.~\ref{app:exp} we provide details about our experimental setup. 
In Sec.~\ref{app:expadd} we present additional figures and tables that accompany those in the main text.

%%%%%%%%%%%%%%%%%%%%%%
\subsection{Local Latent Variable Models}
\label{app:llvm}
%%%%%%%%%%%%%%%%%%%%%%

Another important class of models includes \emph{local} latent variables $\{\bw_n\}$ in addition to
a global latent variable $\bz$, i.e.~models with joint densities of the form:
%%%
\begin{align}
\pth(\DD, \bW, \bz) = \pth(\bz) \prod_{n=1}^N \pth(\bw_n | \bz) \pth(y_n | \bw_n, \bz, \bx_n)
\end{align}
%%%
A viable DAIS-like inference strategy for this class of models that would be scalable to large $N$ is to adopt a \emph{semi-parametric} approach. First, we
introduce a flexible, parametric variational distribution $q_\phi(\bz)$ using, for example, a normalizing flow.
After conditioning on a sample $\bz \sim q_\phi(\bz)$ the $N$ inference problems w.r.t.~$\{\bw_n\}$ effectively decouple.
Consequently we can apply DAIS to each subproblem, resulting in an ELBO that accommodates unbiased mini-batch estimates.

In more detail we proceed as follows.
Introduce a parametric variational distribution $q_\phi(\bz)$ and a mean-field distribution $q_\phi(\bW)$ that factorizes as $q_\phi(\bW)=\prod_n q_\phi(\bw_n)$.
Then write (in this section we suppress momenta terms in the MCMC kernels for brevity)
%%%
\begin{equation}
\begin{split}
    \label{eqn:qfqbllvm}
\qf(\bz, \bW_{0:{K}}) &= q_\phi(\bz) \TT_1(\bW_1 | \bW_0, \bz) \cdots \TT_{K}(\bW_{K} | \bW_{K-1}, \bz) \\
\qb(\bz, \bw_{0:{K}}) &= \pth(\DD, \bW_K, \bz) \tTT_{K}(\bW_{K-1} | \bW_{K}, \bz) \cdots \tTT_1(\bW_0 | \bW_{1}, \bz) \\
    \LL &= \EE_{\qf} \left[ \log \qb - \log \qf \right]
\end{split}
\end{equation}
%%%
where the forward MCMC kernels target a log density of the form (here for $\beta=1$)
%%%
\begin{align}
    \sum_{n=1}^N \left( \log \pth(\bw_n | \bz) + \log \pth(y_n | \bw_n, \bz, \bx_n) \right)
\end{align}
%%%
where $\bz$ is kept fixed throughout HMC dynamics.
Since this density splits up into a sum over $N$ turns, we actually have $N$ independent DAIS chains (assuming our mass matrix is appropriately factorized).
Since the $\log \pth(\DD, \bW_K, \bz)$ term that enters into $\LL$ also splits up into such a sum, this ELBO admits
mini-batch sampling. 
That is, we choose a mini-batch of size $B \ll N$ and run $B$ DAIS chains forward at each optimization step so that e.g.~most
$\{ \bw_n \}$ are not instantiated. This of course is exactly how subsampling works with vanilla parametric mean-field variational distributions.

We can follow the same basic strategy to make this variational distribution more fully DAIS-like \emph{while still supporting subsampling}. The basic
idea is that we will have $N+1$ DAIS chains. The $N$ DAIS chains for each $\bw_n$ will be as above. But we will also introduce a DAIS chain
that is responsible for producing the $\bz$ sample. To enable subsampling this $\bz$-chain must target a surrogate likelihood.

In more detail we proceed as follows. To simplify the equations we assume $K=1$.
%%%
\begin{equation}
\begin{split}
    \label{eqn:qfqbllvmtwo}
    \qf(\tbz_{0:1}, \tbW_{0:1}, \bW_{0:1}) &= q_\phi(\tbz_0, \tbW_0) q_\phi(\bW_0) \TT_{\rm surr}(\tbz_1, \tbW_1 | \tbz_0, \tbW_0)  \TT(\bW_1 | \bW_0, \tbz_1) \\
\qb(\tbz_{0:1}, \tbW_{0:1}, \bW_{0:1}) &= \pth(\DD, \bW_1, \tbz_1) \tTT_{\rm surr}(\tbz_0, \tbW_0 | \tbz_1, \tbW_1)  \tTT(\bW_0 | \bW_1, \tbz_1) \\ 
    \LL &= \EE_{\qf} \left[ \log \qb - \log \qf \right]
\end{split}
\end{equation}
%%%
Here $q_\phi(\tbz_0, \tbW_0)$ and $q_\phi(\bW_0)$ are (distinct) simple parametric variational distributions. $\TT_{\rm surr}$
is a DAIS chain that targets a surrogate likelihood with learnable weights that includes $\Nsurr$ data points and also $\Nsurr$ local latent variables
$\tbW$. As above $\TT(\bW_1 | \bW_0, \tbz_1)$ consists of $N$ independent DAIS chains, all of which are conditioned on $\tbz_1$. 
Note that since these $N$ DAIS chains factorize
there is no need for a surrogate here. We note that the output of the $N+1$ DAIS chains is $\tbz_1, \tbW_1$ \emph{and} $\bW_1 $. $\tbW_1$ is a $\Nsurr$-dimensional `auxiliary variable'
and so we essentially throw it out (nothing depends on it directly)---its only purpose is to `integrate out' $\bW$-uncertainty in the surrogate 
likelihood.\footnote{For this reason we will probably need to pretrain $q_\phi(\tbz_0, \tbW_0)$---perhaps introducing an auxiliary loss function---since $\tbW_1$ will have a weak learning signal.} 
By contrast $\bW_1$ is the `actual' $\bW$ sample that enters into $\pth(\DD, \bW_1, \tbz_1)$, i.e.~our approximate posterior sample is $(\bW_1, \tbz_1)$.
As above we can still subsample the $N$ DAIS chains and get an ELBO
that supports data subsampling. Basically we've replaced the parametric distribution for $\bz$ above with a single DAIS chain that targets a surrogate
likelihood.

Yet another (somewhat simpler)
variant of this approach would forego $\tbW$ transitions in building up the $\bz$ sample and instead parameterize
the surrogate likelihood using a learnable point parameter $\tbW$. 
As should now be evident, the space of models with mixed local/global latent variables opens up lots of possibilities for DAIS-like inference procedures.
For this reason we leave a detailed empirical exploration of the algorithms described above to future work.

%%%%%%%%%%%%%%%%%%%%%%
\subsection{Related Work (extended)}
\label{app:related}
%%%%%%%%%%%%%%%%%%%%%%

In this section we provide an extended discussion of related work.
Many variational objectives that leverage importance sampling (IS) have been proposed.
These include the importance weighted autoencoder (IWAE) \citep{burda2015importance,cremer2017reinterpreting},
the thermodynamic variational objective \citep{masrani2019thermodynamic},
and approaches that make use of Sequential Monte Carlo \citep{le2017auto,maddison2017filtering,naesseth2018variational}.
For a general discussion of IS in variational inference see \citet{domke2018importance}.

As discussed in Sec.~\ref{sec:bg} in the main text our work builds on 
recent work, namely (UHA; \citet{geffner2021mcmc}) and (DAIS; \citet{zhang2021differentiable}),
which can be understood as utilizing unadjusted \emph{underdamped} Langevin steps.
A closely related approach that utilizes unadjusted \emph{overdamped} Langevin steps is described in \cite{thin2021monte}.
We note that earlier attempts to incorporate AIS into a variational framework like \citet{ding2019learning} do
not benefit from a single, unified ELBO-based objective.

An early combination of MCMC methods with variational inference was proposed by \citet{salimans2015markov} and
\citet{wolf2016variational}. A disadvantage of these approaches is the need to learn reverse kernels,
a shortcoming that was later addressed by \citet{caterini2018hamiltonian}. Additional work that explores
combinations of MCMC and variational methods includes \citet{li2017approximate,hoffman2017learning,ruiz2019contrastive}

Bayesian coresets enable users to run MCMC on large datasets after first distilling the data into a smaller
number of weighted data points \citep{huggins2016coresets, bachem2017practical, campbell2018bayesian, campbell2019automated}.
These methods do not support model learning, since they rely on a two-stage approach (i.e.~coreset learning and inference are done separately).
Finally a number of authors have explored MCMC methods that enable data subsampling
\citep{maclaurin2015firefly, quiroz2018speeding, dang2019hamiltonian, zhang2020asymptotically},
including those that leverage stochastic gradients \citep{welling2011bayesian,chen2014stochastic,ma2015complete}.
Finally we note that learning flexible variational distributions parameterized by neural networks, i.e.~normalizing flows,
is another popular black-box approach for improving variational approximations \citep{rezende2015variational}.

%\cite{neal2005hamiltonian,grosse2015sandwiching}

%%%%%%%%%%%%%%%%%%%%%%
\subsection{Computing $\LL_\dais$}
\label{app:daiselbo}
%%%%%%%%%%%%%%%%%%%%%%

We provide a proof of Lemma \ref{lemma:one} in the main text.
We reproduce the argument from \citet{zhang2021differentiable}; see \citet{geffner2021mcmc} for a similar derivation.
The forward kernel $\TT_k$ in Eqn.~\ref{eqn:qfqbdais} can be decomposed as
%%%
\begin{equation}
    \TT_k(\bz_k, \bv_k | \bz_{k-1}, \bv_{k-1}) = 
    \TT_k{^{\rm leap}}(\bz_k, \hat{\bv}_k | \bz_{k-1}, \bv_{k-1})
    \TT_k{^{\rm refresh}}(\bv_{k} | \hat{\bv}_k)
\end{equation}
%%%
Similarly the backward kernel $\tTT_k$ in Eqn.~\ref{eqn:qfqbdais} can be decomposed as
%%%
\begin{equation}
    \tTT_k(\bz_{k-1}, \bv_{k-1} | \bz_k, \bv_k) = 
    \tTT_k{^{\rm refresh}}(\hat{\bv}_k | \bv_k)
    \tTT_k{^{\rm leap}}(\bz_{k-1}, \bv_{k-1} | \bz_k, -\hat{\bv}_k)
\end{equation}
%%%
where we flip the sign of $\hat{\bv}_k$ to account for time reversal.
Since we have 
%%%
\begin{equation}
    \TT_k{^{\rm refresh}}(\bv_{k} | \hat{\bv}_k) = \NN(\bv_{k} | \gamma \hat{\bv}_k, (1-\gamma^2)\mass)        \qquad \qquad
    \tTT_k{^{\rm refresh}}(\hat{\bv}_k | \bv_k) = \NN(\hat{\bv}_k | \gamma \bv_k, (1-\gamma^2)\mass)
\end{equation}
%%%
and
%%%
\begin{equation}
    || \bv_{k} - \gamma \hat{\bv}_k ||^2 - ||\hat{\bv}_k - \gamma \bv_k||^2 = 
    (1-\gamma^2)\left(||\bv_k||^2 - ||\hat{\bv}_k||^2\right)
\end{equation}
%%%
it follows that
%%%
\begin{equation}
    \frac{ \TT_k{^{\rm refresh}}(\bv_{k} | \hat{\bv}_k) }{ \tTT_k{^{\rm refresh}}(\hat{\bv}_k | \bv_k) } = 
    \frac{\NN(\bv_k | \zero, \mass)}{\NN(\hat{\bv}_k | \zero, \mass)}
\end{equation}
%%%
Furthermore, since the leapfrog step is volume preserving and reversible we also have that
%%%
\begin{equation}
    \label{app:equaldensity}
    \TT_k{^{\rm leap}}(\bz_k, \hat{\bv}_k | \bz_{k-1}, \bv_{k-1}) = 
    \tTT_k{^{\rm leap}}(\bz_{k-1}, \bv_{k-1} | \bz_k, -\hat{\bv}_k)
\end{equation}
%%%
From these equations the only tricky part of Lemma \ref{lemma:one}, i.e.~the derivation of the kinetic
energy difference correction, follows.
Importantly, Eqn.~\ref{app:equaldensity} is true regardless of the target density used to define 
$\TT_k{^{\rm leap}}$ and $\tTT_k{^{\rm leap}}$; in particular using a surrogate log likelihood leaves Lemma \ref{lemma:one} intact.

%%%%%%%%%%%%%%%%%%%%%%
\subsection{General properties of NS-DAIS and SL-DAIS}
\label{app:nssldais}
%%%%%%%%%%%%%%%%%%%%%%

%%%
We prove Proposition~\ref{prop:nssl} in the main text.
In particular we show that the NS-DAIS and SL-DAIS approximate posteriors $\qf(\bz_K)$ satisfy the inequality
%%%
\begin{equation}
\begin{split}
    \log \pth(\DD) \ge \LL + \KL( \qf(\bz_{K}) | \pth(\bz_K | \DD) )
\end{split}
\end{equation}
%%%
where $\LL$ is $\LL_\nsdais$ or $\LL_\sldais$.
First consider SL-DAIS
and let $\tqb(\bz_{0:K}, \bv_{0:K})$ be the normalized distribution corresponding to $\qb(\bz_{0:K}, \bv_{0:K})$ so
that $\log \qb(\bz_{0:K}, \bv_{0:K}) = \log \pth(\DD) + \log \tqb(\bz_{0:K}, \bv_{0:K})$ and write
%%%
\begin{align}
    \log \pth(\DD) &= \EE_{\qf} \Big[ \log \qb(\bz_{0:K}, \bv_{0:K}) \!-\! \log  \qf(\bz_{0:K}, \bv_{0:K}) \Big] + \\
                         &\qquad\KL( \qf(\bz_{0:K}, \bv_{0:K}) | \tqb(\bz_{0:K}, \bv_{0:K}) )  \nonumber \\
                   &= \LL_\sldais + \KL( \qf(\bz_{0:K}, \bv_{0:K}) | \tqb(\bz_{0:K}, \bv_{0:K}) )  \\
                   &= \LL_\sldais + \KL( \qf(\bz_{K}) | \pth(\bz_K | \DD) ) \;+ \label{eqn:sldaisklproof}\\
                       &\qquad\;\;\EE_{\qf(\bz_K) } \left[ \KL( \qf(\bz_{0:K}, \bv_{0:K} | \bz_K) | \tqb(\bz_{0:K}, \bv_{0:K} | \bz_K) ) \right] \nonumber 
\end{align}
%%%
where we appealed to the chain rule of KL divergences which reads
%%%
\begin{equation}
\KL( q(a, b) | p(a, b) ) = \KL( q(a) | p(a) ) + \EE_{q(b)} \left[ \KL(q(a | b) | p(a | b) ) \right]
\end{equation}
%%%
The result then follows since the second KL divergence in Eqn.~\ref{eqn:sldaisklproof} is non-negative.
The result for NS-DAIS follows by the same argument, with the difference that $\JJ$
is now one of the `auxiliary' variables $\{ \bz_{0:K}, \bv_{0:K} \}$.
Note that the $\II$ index plays no role in this proof, since its role is to facilitate 
unbiased mini-batch estimates, i.e.~it has no impact on the value of $\LL_\nsdais$ or $\LL_\sldais$
(see the next section for details).

%%%%%%%%%%%%%%%%%%%%%%
\subsection{NS-DAIS}
\label{app:nsdais}
%%%%%%%%%%%%%%%%%%%%%%

We first elaborate how NS-DAIS is formulated.
The forward and backward kernels are defined as:
%%%
\begin{equation}
\begin{split}
    \label{eqn:appqfqbnsdais}
    &\qf(\bz_{0:K}, \bv_{0:K},\II, \JJ)
    = q(\II)q(\JJ) q_0(\bz_0) q_0(\bv_0) \textstyle{\prod}_{k=1}^K \TT_k(\bz_k, \bv_k | \bz_{k-1}, \bv_{k-1}, \JJ) \\
    &\qb(\bz_{0:K}, \bv_{0:K}, \II, \JJ) = q(\II)q(\JJ) \pth( \bz_K | \DD_{\II})^{N/B} \pth(\bz_K) 
    \textstyle{\prod}_{k=1}^K \tTT_k(\bz_{k-1}, \bv_{k-1} | \bz_k, \bv_k, \JJ) \nonumber
\end{split}
\end{equation}
%%%
where $\JJ$ controls the potential energy used to guide the HMC dynamics
and the corresponding ELBO is given by
%%%
\begin{equation}
\begin{split}
    \LL_\nsdais &\equiv \EE_{\qf} \Big[ \log \qb(\bz_{0:K}, \bv_{0:K}, \II, \JJ) \!-\! \log  \qf(\bz_{0:K}, \bv_{0:K}, \II, \JJ) \Big] \\
    &= \EE_{\qf^\prime} \Big[ \log \qb^\prime(\bz_{0:K}, \bv_{0:K}, \JJ) \!-\! \log  \qf^\prime(\bz_{0:K}, \bv_{0:K}, \JJ) \Big]
\end{split}
\end{equation}
%%%
where
%%%
\begin{align}
    &\qf^\prime(\bz_{0:K}, \bv_{0:K}, \JJ)
    = q(\JJ) q_0(\bz_0) q_0(\bv_0) \textstyle{\prod}_{k=1}^K \TT_k(\bz_k, \bv_k | \bz_{k-1}, \bv_{k-1}, \JJ) \\
    &\qb^\prime(\bz_{0:K}, \bv_{0:K}, \JJ) = q(\JJ) \pth( \bz_K | \DD) \pth(\bz_K) 
    \textstyle{\prod}_{k=1}^K \tTT_k(\bz_{k-1}, \bv_{k-1} | \bz_k, \bv_k, \JJ) \nonumber
\end{align}
%%%
are the forward and backward kernels without the $\II$ mini-batch index.\footnote{Note that these are the
kernels that were used in the proof in Sec.~\ref{app:nssldais}.}
Note that here and elsewhere $q_0(\bv_0)=\NN(\bv_0 | \zero, \mass)$ is the momentum distribution.
To be more specific, for each $\JJ$ the potential energy that guides each $\TT_k$ is given by
%%%
\begin{align}
    \label{eqn:vk}
    V_k(\bz | \JJ) = -(1-\beta_k) \log q_0(\bz) - \beta_k \left( \Psi_0(\bz) + \tfrac{N}{B}\PsiL(\DD_\JJ, \bz) \right)
\end{align}
%%%
See Algorithm~\ref{algo:nsdais} for the complete procedure
and see Algorithm~\ref{algo:dais} to make a direct comparison to DAIS. 

\begin{algorithm}[tb]
    \caption{NS-DAIS: Naive Subsampling Differentiable Annealed Importance Sampling. We
    highlight in \highlight{\textrm{blue}} where the algorithm differs from DAIS. 
    To recover DAIS we substitute $\frac{N}{B}\PsiL(\DD_\JJ, \bz) \rightarrow \PsiL(\DD, \bz)$,
    $\frac{N}{B}\PsiL(\DD_\II, \bz_K) \rightarrow \PsiL(\DD, \bz_K)$, and
    $B\rightarrow N$ (or in other words remove all mini-batch sampling). 
    }
   \label{algo:nsdais}
%%%%%%%%%%%%%%%
\begin{algorithmic}
    \STATE {\bfseries Input:} model log prior density $\Psi_0(\bz)$,
                              model log likelihood $\PsiL(\DD, \bz)$, 
                              base variational distribution $q_0(\bz)$, number of steps $K$,
                              inverse temperatures $\{ \beta_k \}$, 
                              step size $\eta$, momentum refresh parameter $\gamma$, 
                              mass matrix $\mass$,
                              dataset $\DD$ of size $N$,
                              mini-batch size $B$
    \STATE {\bfseries Initialize:}  $\bz_0 \sim q_0$, $\bv_0 \sim \NN(\zero, \mass)$, $\LL \leftarrow -\log q_0 (\bz_0)$ \\
    \highlight{\textrm{Sample}} mini-batch indices $\JJ \! \subset \! \{1, ..., N\}$ with $|\JJ|=B$
    %%%%
    \FOR{$k=1$ {\bfseries to} $K$}
    %%%
    \STATE $\hat{\bz}_k \leftarrow \bz_{k-1} + \frac{\eta}{2}\mass^{-1}\bv_{k-1}$ 
    \STATE $\bg_k \leftarrow \! \nabla_{\!\bz} \! \left\{\!\beta_k \!\left( \! \Psi_0(\bz) \!+\! 
                                               \highlight{\tfrac{N}{B}\PsiL(\DD_\JJ, \bz)} \! \right) \!+\! 
                                               (1 \!-\! \beta_k) \log q_0(\bz)\right\}\!\bigg\rvert_{\bz=\hat{\bz}_k} $
    \STATE $\hat{\bv}_k \leftarrow \bv_{k-1} + \eta \bg_k $
    \STATE $\bz_{k} \leftarrow \hat{\bz}_k+ \frac{\eta}{2}\mass^{-1}\hat{\bv}_{k}$
    \vspace{1mm}
    \IF{$k < K$}
        \STATE $\bv_k \leftarrow \gamma \hat{\bv}_k + \sqrt{1 - \gamma^2}\eps, \; \eps \sim \NN(\zero, \mass)$
    \ENDIF
    \STATE $\LL \leftarrow \LL + \log \NN(\hat{\bv}_k, \mass) - \log \NN(\bv_{k-1}, \mass)$
    %%%
    \ENDFOR \\
    \highlight{\textrm{Sample}} mini-batch indices $\II \! \subset \! \{1, ..., N\}$ with $|\II|=B$
    \STATE {\bfseries Return:} $\LL \mathrel{+} \Psi_0(\bz_K) + \highlight{\tfrac{N}{B} \PsiL(\DD_{\II}, \bz_K)} $ 
    %%%
\end{algorithmic}
\end{algorithm}

\begin{algorithm}[tb]
    \caption{DAIS: Differentiable Annealed Importance Sampling.
    We provide a complete description of DAIS \citep{geffner2021mcmc, zhang2021differentiable} in our notation.
    }
   \label{algo:dais}
%%%%%%%%%%%%%%%
\begin{algorithmic}
    \STATE {\bfseries Input:} model log prior density $\Psi_0(\bz)$,
                              model log likelihood $\PsiL(\DD, \bz)$, 
                              base variational distribution $q_0(\bz)$, number of steps $K$,
                              inverse temperatures $\{ \beta_k \}$, 
                              step size $\eta$, momentum refresh parameter $\gamma$, 
                              mass matrix $\mass$,
    \STATE {\bfseries Initialize:}  $\bz_0 \sim q_0$, $\bv_0 \sim \NN(\zero, \mass)$, $\LL \leftarrow -\log q_0 (\bz_0)$ \\
    %%%%
    \FOR{$k=1$ {\bfseries to} $K$}
    %%%
    \STATE $\hat{\bz}_k \leftarrow \bz_{k-1} + \frac{\eta}{2}\mass^{-1}\bv_{k-1}$ 
    \STATE $\bg_k \leftarrow \! \nabla_{\!\bz} \! \left\{\!\beta_k \!\left( \! \Psi_0(\bz) \!+\! 
                                               \PsiL(\DD, \bz) \! \right) \!+\! 
                                               (1 \!-\! \beta_k) \log q_0(\bz)\right\}\!\bigg\rvert_{\bz=\hat{\bz}_k} $
    \STATE $\hat{\bv}_k \leftarrow \bv_{k-1} + \eta \bg_k $
    \STATE $\bz_{k} \leftarrow \hat{\bz}_k+ \frac{\eta}{2}\mass^{-1}\hat{\bv}_{k}$
    \vspace{1mm}
    \IF{$k < K$}
        \STATE $\bv_k \leftarrow \gamma \hat{\bv}_k + \sqrt{1 - \gamma^2}\eps, \; \eps \sim \NN(\zero, \mass)$
    \ENDIF
    \STATE $\LL \leftarrow \LL + \log \NN(\hat{\bv}_k, \mass) - \log \NN(\bv_{k-1}, \mass)$
    %%%
    \ENDFOR \\
    \STATE {\bfseries Return:} $\LL \mathrel{+} \Psi_0(\bz_K) + \PsiL(\DD, \bz_K) $ 
    %%%
\end{algorithmic}
\end{algorithm}

Now we prove Proposition \ref{prop:nsdais} in the main text.
As in the main text let $q(\JJ)$ denote the distribution that corresponds to 
sampling mini-batches of $B$ indices $\JJ \subset \{1, ..., N \}$ without
replacement. Let $\pth(\bz | \DD_\JJ)$ denote the posterior that corresponds
to the data subset $\DD_\JJ$ with appropriately scaled likelihood term, i.e.
%%%
\begin{equation}
    \pth(\bz | \DD_\JJ) \propto \pth(\DD_\JJ | \bz)^{N/B} \pth(\bz)
\end{equation}
%%%
Further denote the `aggregate pseudo-posterior' by 
%%%
\begin{equation}
    \pth^\agg(\bz | \DD) \equiv \EE_{q(\JJ)} \left[ \pth(\bz | \DD_\JJ) \right]
\end{equation}
%%%
We have (appealing to the convexity of the KL divergence)
%%%
\begin{align}
    \KL( \qf(\bz_K ) | \pth^\agg(\bz | \DD) ) 
    &= \KL( \EE_{q(\JJ)} \left[ \qf(\bz_K | \JJ ) \right]| \EE_{q(\JJ)} \left[ \pth(\bz_K | \DD_\JJ) \right] )  \\
    &\le \EE_{q(\JJ)} \left[ \KL( \qf(\bz_K | \JJ) | \pth(\bz_K | \DD_\JJ) ) \right] \\
    &=  \OO(K^{-1/2})
\end{align}
%%%
where in the last line we used that $q(\JJ)$ has finite support and where we have appealed
to the convergence result in \citet{zhang2021differentiable} to conclude that for each $\JJ$
we have that $\KL( \qf(\bz_K | \JJ) | \pth(\bz_K | \DD_\JJ) )  =  \OO(K^{-1/2})$.
We note that, as is well known, convergence in KL divergence implies convergence w.r.t.~the total variation distance.

%%%%%%%%%%%%%%%%%%%%%%
\subsection{SL-DAIS}
\label{app:sldais}
%%%%%%%%%%%%%%%%%%%%%%

Before we turn to a proof of Proposition~\ref{prop:sldais} in the main text, we give a more formal
description of SL-DAIS.
The forward and backward kernels are defined as:
%%%
\begin{equation}
\begin{split}
    \label{eqn:appqfqbsldais}
    &\qf(\bz_{0:K}, \bv_{0:K}, \II) = q(\II) q_0(\bz_0) q_0(\bv_0) \textstyle{\prod}_{k=1}^K \TT_k(\bz_k, \bv_k | \bz_{k-1}, \bv_{k-1}, \PsiLhat) \\
    &\qb(\bz_{0:K}, \bv_{0:K}, \II) = q(\II) \pth( \bz_K | \DD_{\II})^{N/B} \pth(\bz_K) 
    \textstyle{\prod}_{k=1}^K \tTT_k(\bz_{k-1}, \bv_{k-1} | \bz_k, \bv_k, \PsiLhat) \nonumber
\end{split}
\end{equation}
%%%
and the corresponding ELBO is given by
%%%
\begin{equation}
\begin{split}
    \LL_\sldais &\equiv \EE_{\qf} \Big[ \log \qb(\bz_{0:K}, \bv_{0:K}, \II) \!-\! \log  \qf(\bz_{0:K}, \bv_{0:K}, \II) \Big] \\
    &= \EE_{\qf^\prime} \Big[ \log \qb^\prime(\bz_{0:K}, \bv_{0:K}) \!-\! \log  \qf(\bz_{0:K}, \bv_{0:K}) \Big]
\end{split}
\end{equation}
%%%
where
%%%
\begin{equation}
\begin{split}
    &\qf^\prime(\bz_{0:K}, \bv_{0:K}) = q_0(\bz_0) q_0(\bv_0) \textstyle{\prod}_{k=1}^K \TT_k(\bz_k, \bv_k | \bz_{k-1}, \bv_{k-1}, \PsiLhat) \\
    &\qb^\prime(\bz_{0:K}, \bv_{0:K}) = \pth( \bz_K | \DD) \pth(\bz_K) 
    \textstyle{\prod}_{k=1}^K \tTT_k(\bz_{k-1}, \bv_{k-1} | \bz_k, \bv_k, \PsiLhat) \nonumber
\end{split}
\end{equation}
%%%
are the forward and backward kernels without the $\II$ mini-batch index.

We now provide a proof of Proposition~\ref{prop:sldais} in the main text.
First we restate some of the equations and notation from the main text.
We consider Bayesian linear regression with a prior $\pth(\bz) = \NN(\bmu_0, \bLam_0^{-1})$ and a 
likelihood $\prod_n \NN(y_n | \bz \cdot \bx_n, \sigmaobs^2)$, where each $y_n \in \RR$ and $\bx_n \in \RR^D$.
The exact posterior is given by
$\NN(\bmu_\post, \bLam_\post^{-1})$ where $\bmu_\post = \bLam_\post^{-1} (\bLam_0 \bmu_0 + \frac{1}{\sigmaobs^2} \bX^\top \by)$
and $\bLam_\post = \bLam_0 + \frac{1}{\sigmaobs^2} \bX^\top \bX$.
Throughout we work under the following set of simplifying assumptions:
%%%
\addtocounter{ass}{-1}
\begin{restatable}{ass}{assumption}
\label{assapp}
We use full momentum refreshment ($\gamma = 0$), 
equally spaced inverse temperatures $\{ \beta_k \}$,
a step size that varies as $\eta \sim K^{-1/4}$,
and the prior as the base distribution (i.e.~$q_0(\bz) \!\rightarrow\! \pth(\bz)$).
\end{restatable}

Next we define the parameters of the `surrogate posterior'
$\NN(\tbmu_\post, \tbLam_\post)$ that corresponds to the surrogate log likelihood
$\PsiLhat(\bz)$ specified by $\dbA$ and $\dbB$, where $\dbA$ and $\dbB$ control the degree to which 
$\PsiLhat(\bz)$ is incorrectly specified. Given our parameterization
%%%
\begin{equation}
    \nabla_\bz \PsiLhat(\bz) =  \bA + \dbA - (\bB + \dbB) \bz 
\end{equation}
%%%
it is easy to verify that
%%%
\begin{equation}
\begin{split}
\tbLam_\post = \bLam_0 + \bB +  \dbB = \bLam_\post + \dbB  \\
\tbmu_\post = \tbLam_\post^{-1}  \left( \bLam_0 \mu_0 +  \bA + \dbA\right) 
=  \tbLam_\post^{-1}  \left( \bLam_\post \bmu_\post + \dbA \right)
\end{split}
\end{equation}
%%%
Expanding these expressions in powers of $\dbA$ and $\dbB$ we have:
%%%
\begin{equation}
\begin{split}
\tbLam_\post^{-1} \approx \bLam_\post^{-1} -  \bLam_\post^{-1} \dbB \bLam_\post^{-1} \\
\tbmu_\post \approx \bmu_\post  + \bLam_\post^{-1} \dbA - \bLam_\post^{-1} \dbB \bmu_\post 
\end{split}
\label{surr_post_approx}
\end{equation}
%%%
These expressions also imply that
%%%
\begin{equation}
\det \tbLam_\post = \det \bLam_\post \det \left( \mathbb{1}_D - \bLam_\post^{-1} \dbB \right ) 
                       \approx \det \bLam_\post \left(1 + \trace \bLam_\post^{-1} \dbB \right)
\end{equation}
%%%
which yields
%%%
\begin{equation}
    \label{eqn:logdet}
\log |\bLam_\post^{-1} \tbLam_\post | \approx \trace  \bLam_\post^{-1} \dbB .
\end{equation}

Next we restate some results from Sec.~4.1 and the supplementary materials in \citet{zhang2021differentiable}, suitably adapting them to our setting and notation.
Where appropriate we use subscripts/superscripts to distinguish DAIS and SL-DAIS terms.
%%%
\begin{restatable}{lemma}{lemmatwo}
    In the case of Bayesian linear regression under Assumption~\ref{ass} each of the $K+1$ random variables $\{ \bz_k \}$ that enter DAIS are normally distributed as $\bz_k \sim \NN(\bmu_k, \bLam_k^{-1})$.
    If we run DAIS for $K$ steps the ELBO gap is given by 
%%%%
	\begin{equation}
		\begin{split}
		   & \log p(\DD) -   \LL_\dais  = \\
		   &\qquad  \underbrace{\frac{1}{2} \| \bmu_K - \bmu_\post \|_{\bLam_\post}^2}_{\rm (I)}
		   +\underbrace{\frac{1}{2} \trace (\bLam_\post \bLam_K^{-1}) - \frac{D}{2}}_{\rm (II)}
		   + \underbrace{\frac{1}{2} \log |\bLam_\post^{-1}\bLam_0| - \EE_{q_\text{fwd}}\left[ \sum_{k=1}^K \log \frac{\NN(\hat{\bv}_k, \mass)}{\NN(\bv_{k-1}, \mass)}  \right]}_{\rm (III)}
		\end{split}
	\label{bound_diff}
	\end{equation}
	\label{lemma:two}
    where $||\bmu_K - \bmu_\post||_{\bLam_\post}^2 = (\bmu_K - \bmu_\post)^{\rm T} \bLam_\post (\bmu_K - \bmu_\post) $.
    Moreover, as is evident from the derivation in Sec.~B.2 in \citet{zhang2021differentiable}, 
    this equation also holds for the gap $\log p(\DD) - \LL_\sldais$ if
    we substitute the SL-DAIS forward kernel $q_\text{fwd}$ into the expectation in (III)
    and $\bmu_k$ and $\bLam_k$ are defined in terms of the $\{ \bz_k \}$ that enter SL-DAIS. 
%%%
    Additionally for SL-DAIS\footnote{This is evident from the proof of Lemma 1 in \citet{zhang2021differentiable}. The analogous result is also true for DAIS but we do not require this for the proof.} we have that
%%%
	\begin{equation}
        \label{eqn:lemmatwo}
		\begin{split}
			\| \bmu_K^\sldais - \tbmu_\post \| = \OO(K^{-1/2}) \qquad \qquad 
            \| (\bLam_K^\sldais)^{-1} - \tbLam_\post^{-1} \| = \OO(K^{-1/2}) \\
			\frac{1}{2} \log |\tbLam_\post^{-1}\bLam_0| - \EE_{q_\text{fwd}}\left[ \sum_{k=1}^K \log \frac{\NN(\hat{\bv}_k, \mass)}{\NN(\bv_{k-1}, \mass)}  \right] =  \OO(K^{-1/2})
		\end{split}
	\end{equation}
\end{restatable}
%%%
With these ingredients we can now proceed with the proof. 
From Eqn.~\ref{surr_post_approx} and Eqn.~\ref{eqn:lemmatwo} we have 
%%%
\begin{equation}
	\| \bmu_K^\sldais - \bmu_\post \| \le \| \bmu_K^\sldais - \tbmu_\post \| + \| \tbmu_\post - \bmu_\post \| =  \OO(K^{-1/2}) + \OO(\|\dbA\| )+ \OO(\|\dbB\|).
\end{equation}
%%%
Hence for SL-DAIS we have that
%%%
\begin{equation}
	{\rm (I)}  = \OO(K^{-1 / 2}) + \OO(\|\dbA\|^2 + \|\dbB\|^2).
\end{equation}
%%%
Once again appealing to Eqn.~\ref{surr_post_approx} and Eqn.~\ref{eqn:lemmatwo} we have
%%%
\begin{equation}
	\begin{aligned}
        \trace (\bLam_\post (\bLam_K^\sldais)^{-1}) - D &= \trace (\bLam_\post( (\bLam_K^\sldais)^{-1} - \bLam_\post^{-1} ))\\
        &= \trace (\bLam_\post ((\bLam_K^\sldais)^{-1} -\tbLam_\post^{-1} ) ) + \trace (\bLam_\post (\tbLam_\post^{-1}-\bLam_\post^{-1} ) ) \\
		&\approx \OO(K^{-1/2}) + \trace  \bLam_\post^{-1} \dbB
	\end{aligned}
\end{equation}
%%%
so that we get the following estimate for the second term of Eqn.~\ref{bound_diff}:
%%%
\begin{equation}
	{\rm (II)} = \OO(K^{-1/2})  + \frac{1}{2} \trace  \bLam_\post^{-1} \dbB
\end{equation}
%%%
Finally, from Eqn.~\ref{eqn:logdet} and Eqn.~\ref{eqn:lemmatwo} we have
\begin{equation}
	{\rm (III)} = \frac{1}{2}\log |\bLam_\post^{-1} \tbLam_\post | + \OO(K^{-1/2}) =  \frac{1}{2}\trace  \bLam_\post^{-1} \dbB +  \OO(K^{-1/2})
\end{equation}
which ends the first part of the proof for Proposition~\ref{prop:sldais}.
The fact that the KL divergence between $\qf(\bz_K)$ and the posterior $\pth(\bz_K | \DD)$ is bounded
by the same quantity easily follows from standard arguments, see e.g.~the proof of Proposition~\ref{prop:nssl}
in Sec.~\ref{app:nssldais}.

%%%%%%%%%%%%%%%%%%%%%%
\subsection{Experimental details}
\label{app:exp}
%%%%%%%%%%%%%%%%%%%%%%

All experiments use $64$-bit floating point precision.
We use the Adam optimizer with default momentum hyperparameters in all experiments \citep{kingma2014adam}.
In all optimization runs we do $3 \times 10^5$ optimization steps with an initial learning
rate of $10^{-3}$ that drops to $10^{-4}$ and $10^{-5}$ at iteration $10^5$ and $2 \times 10^5$, respectively.
Similar to \citet{geffner2021mcmc} we parameterize the step size $\eta_k$ in the $k^{\rm th}$ iteration of
DAIS/NS-DAIS/SL-DAIS as
%%%
\begin{equation}
    \eta_k = {\rm clip}(\tilde{\eta} + \kappa \beta_k, {\rm min}=0, {\rm max}=\eta_{\rm max})
\end{equation}
%%%
where $\tilde{\eta}$ and $\kappa$ are learnable parameters and we choose $\eta_{\rm max}=0.25$.
The inverse temperatures $\{ \beta_k \}$ are parameterized using the exponential transform
to enforce positivity and a cumulative summation to enforce monotonicity.
For SL-DAIS the learnable weights $\bomega$ are uniformly initialized so that their total
weight is equal to the number of data points, i.e.~$\sum_n \omega_n = N$.
In all experiments we use a diagonal mass matrix $\mass$. We initialize $\tilde{\eta}$ to be
small, e.g.~$\tilde{\eta}\sim10^{-4} - 10^{-2}$, and initialize $\kappa$ to $\kappa=0$.
We initialize $\gamma$ to $\gamma=0.9$.
Unless noted otherwise (see GP models), we learn all parameters that define the model and 
variational distribution jointly in a single optimization run, 
i.e.~without any kind of pre-training (in contrast to \citet{geffner2021mcmc}).
We found that this generally worked better, although in a few cases pre-training led to better results
when the base distribution was a multivariate Normal.

%%%%%%%%%%%%%%%%%%%%%%
\subsubsection{Surrogate log likelihoods}
\label{app:ablation}
%%%%%%%%%%%%%%%%%%%%%%

All methods use $K=8$. We use 50k MC samples to estimate ELBOs after training.
The training set has 50k data points and all replications are for the same training set.
We consider eight different strategies for defining a surrogate log likelihood in the case of logistic regression:
%%%
\begin{enumerate}
\item \texttt{RAND}: 
    We randomly choose $\Nsurr$ surrogate data points $\{ (\ty_n, \tbx_n) \} \subset \DD$, introduce
    a $\Nsurr$-dimensional vector of learnable weights $\bomega$, and let $\PsiLhat(\bz) = \sum_n \omega_n \log p(\ty_n | \bz, \tbx_n)$.
\item \texttt{RAND$\pm$}:
    We randomly choose $\Nsurr$ surrogate covariates $\{\tbx_n\}$ and
    introduce two $\Nsurr$-dimensional vectors of learnable weights, $\bomega^{-}$ and $\bomega^{+}$. We then write $\PsiLhat(\bz) = \sum_n \omega_n^{+} \log p(y_n=1 | \bz, \tbx_n) + \sum_n \omega_n^{-} \log p(y_n=0 | \bz, \tbx_n)$.
\item \texttt{CS-INIT}: 
    We use a Bayesian coreset algorithm \citep{huggins2016coresets} to
    choose $\Nsurr$ surrogate data points, introduce
    a $\Nsurr$-dimensional vector of learnable weights $\bomega$, and let $\PsiLhat(\bz) = \sum_n \omega_n \log p(\ty_n | \bz, \tbx_n)$.
\item \texttt{CS-INIT$\pm$}:
    We proceed as in \texttt{CS-INIT} but ignore the $\{\ty_n\}$ returned by the coreset algorithm and instead
    introduce two $\Nsurr$-dimensional vectors of learnable weights, $\bomega^{-}$ and $\bomega^{+}$. We then write $\PsiLhat(\bz) = \sum_n \omega_n^{+} \log p(y_n=1 | \bz, \tbx_n) + \sum_n \omega_n^{-} \log p(y_n=0 | \bz, \tbx_n)$.
\item \texttt{CS-FIX}: 
    As in \texttt{CS-INIT} we use a coreset algorithm to choose the surrogate data points but instead of learning $\bomega$
    we use the weights provided by the coreset algorithm.
\item \texttt{KM-INIT$\pm$}: 
    We use k-means++ to extract $\Nsurr$ cluster centroids from the set of $N$ covariates in $\DD$ and
    introduce two $\Nsurr$-dimensional vectors of learnable weights, $\bomega^{-}$ and $\bomega^{+}$. 
    We then write $\PsiLhat(\bz) = \sum_n \omega_n^{+} \log p(y_n=1 | \bz, \tbx_n) + \sum_n \omega_n^{-} \log p(y_n=0 | \bz, \tbx_n)$
    and treat the surrogate covariates $\{\tbx_n\}$ as learnable parameters.
\item \texttt{KM-FIX$\pm$}: 
    We proceed as in \texttt{KM-INIT$\pm$} except the surrogate covariates returned by the clustering algorithm
    remain fixed.
\item \texttt{NN}:
    We parameterize $\PsiLhat(\bz)$ as a neural network with two hidden layers, ELU non-linearities, and 100 neurons per hidden layer.
\end{enumerate}
%%%

\begin{table*}[h]
    \centering
\begin{tabular}{l|l|l|l|l|l|l}
\hline
{\small Dataset} & \multicolumn{3}{l|}{{\small Higgs}}    & \multicolumn{3}{l}{{\small SUSY}} \\ \hline
{\small $\Nsurr$}     & {\small $64$}   & {\small $256$}  & {\small $1024$} & {\small $64$}  & {\small $256$} & {\small $1024$} \\ \hline \hline
\small{\texttt{RAND}}       & \uShiggsRAND    & \uMhiggsRAND    & \uLhiggsRAND    & \uSsusyRAND    & \uMsusyRAND    & \uLsusyRAND    \\ \hline 
\small{\texttt{RAND$\pm$}}  & \uShiggsRANDpm  & \uMhiggsRANDpm  & \uLhiggsRANDpm  & \uSsusyRANDpm  & \uMsusyRANDpm  & \uLsusyRANDpm  \\ \hline 
\small{\texttt{CS-INIT}}    & \uShiggsCS      & \uMhiggsCS      & \uLhiggsCS      & \uSsusyCS      & \uMsusyCS      & \uLsusyCS      \\ \hline
\small{\texttt{CS-INIT$\pm$}} & \uShiggsCSpm  & \uMhiggsCSpm    & \uLhiggsCSpm    & \uSsusyCSpm    & \uMsusyCSpm    & \uLsusyCSpm    \\ \hline
\small{\texttt{CS-FIX}}     & \uShiggsFIXCS   & \uMhiggsFIXCS   & \uLhiggsFIXCS   & \uSsusyFIXCS   & \uMsusyFIXCS   & \uLsusyFIXCS   \\ \hline
\small{\texttt{KM-INIT$\pm$}} & \uShiggsKMpm  & \uMhiggsKMpm    & \uLhiggsKMpm    & \uSsusyKMpm    & \uMsusyKMpm    & \uLsusyKMpm    \\ \hline
\small{\texttt{KM-FIX$\pm$}} & \uShiggsFIXKMpm & \uMhiggsFIXKMpm & \uLhiggsFIXKMpm & \uSsusyFIXKMpm & \uMsusyFIXKMpm & \uLsusyFIXKMpm \\ \hline
\small{\texttt{NN}}         & \multicolumn{3}{l|}{\uShiggsNN}   & \multicolumn{3}{l}{\uSsusyNN}  \\ \hline
\end{tabular}
\caption{This table is an expanded version of Table~\ref{table:ablation} in Sec~\ref{sec:ablation}. We compare eight different parameterizations for the surrogate log likelihood $\PsiLhat$.
We consider two logistic regression datasets.
For the seven strategies that make use of surrogate data points, we vary $\Nsurr \in \{64, 256, 1024\}$.
We report ELBO improvements (in nats) above a mean-field Normal baseline.
See Sec.~\ref{sec:ablation}.}
\label{table:ablation_full}
\end{table*}

%%%%%%%%%%%%%%%%%%%%%%
\subsubsection{Classifying imbalanced data}
\label{app:imbalanced}
%%%%%%%%%%%%%%%%%%%%%%

We set $\Nsurr=B=256$ and $K=8$. We use the SUSY dataset with 20k data points held out for testing.
The rare class corresponds to signal events.
For each ratio of rare to non-rare class we keep the dataset fixed so that replications only differ with respect
to the random number seed that controls optimization.

%%%%%%%%%%%%%%%%%%%%%%
\subsubsection{Logistic regression}
\label{app:logistic}
%%%%%%%%%%%%%%%%%%%%%%

When running HMCECS \citep{dang2019hamiltonian}, we use the `perturbed' approach with a NUTS kernel and 
use a control variate based on a second-order Taylor approximation around the maximum a posteriori estimate. 
In addition, to facilitate mixing we use block pseudo-marginal updates for data subsample indices with 25 blocks
\citep{tran2016block}.
We run for a total of $15 \times 10^3$ iterations, discard the first 
$5 \times 10^3$ samples for burn-in, thin the remaining $10 \times 10^3$ samples by a factor of 10, 
and use these $10^3$ thinned samples to compute posterior predictive distributions on held-out test data.
For NUTS we use the default settings in NumPyro, in particular a diagonal mass matrix that is adapted
during warm-up and a maximum tree depth of $10$.

For NUTS-CS we identify $1024$ coresets using the coreset algorithm described in \citet{huggins2016coresets}.
Note that in most cases the total number of coresets returned by the algorithm is somewhat less than $1024$,
e.g.~$1009$ or $1017$. We use the settings recommended in \citet{huggins2016coresets}.
In particular we use kmeans++ to construct a clustering with $k=6$ clusters.
We set the hyperparameter $a$ to $a=6$. 
We then run NUTS \citet{hoffman2014no} on a logistic regression model conditioned on the weighted coresets using the same number of NUTS iterations, mass matrix, etc.,~as for HMCECS.

The Higgs and SUSY datasets are described in \citep{baldi2014searching} and available from the UCI repository \citep{asuncion2007uci}.
The MiniBooNE dataset is from \citep{roe2005boosted} and is likewise available from UCI.
For the two CovType datasets \citep{blackard1999comparative}, which are also vailable from UCI,
we reduce the 7-way classification problem to two binary classifications
problems: i) fir versus rest; and ii) pine versus rest. 
%    if binary and not fir:
%        y = np.bitwise_or(y == 2, y == 3)
%    elif binary and fir:
%        y = np.bitwise_or(y == 1, y == 6)
%  1 -- Spruce/Fir
%  2 -- Lodgepole Pine
%  3 -- Ponderosa Pine
%  4 -- Cottonwood/Willow
%  5 -- Aspen
%  6 -- Douglas-fir
%  7 -- Krummholz

%%%
\begin{figure*}[ht]
\begin{center}
\includegraphics[width=.95\columnwidth]{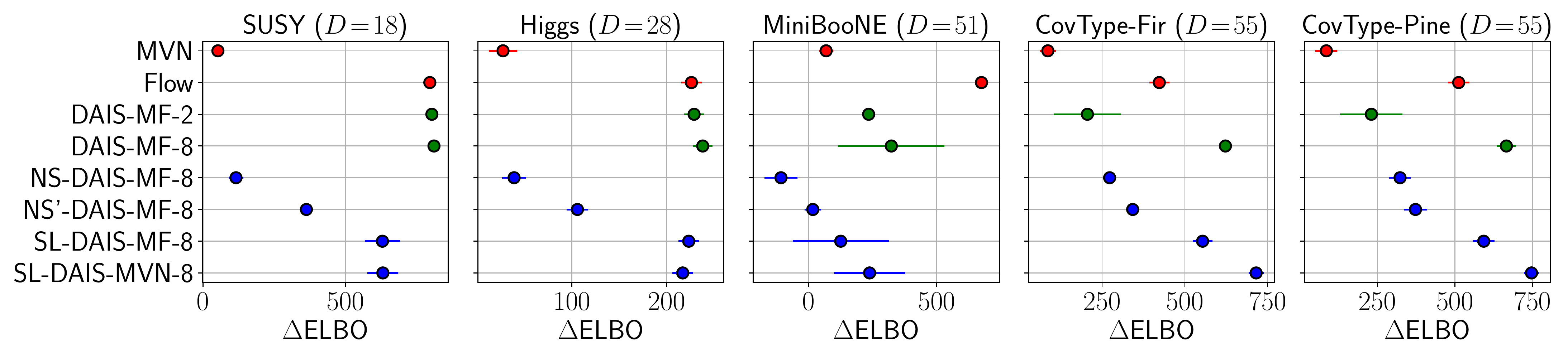}
\includegraphics[width=.95\columnwidth]{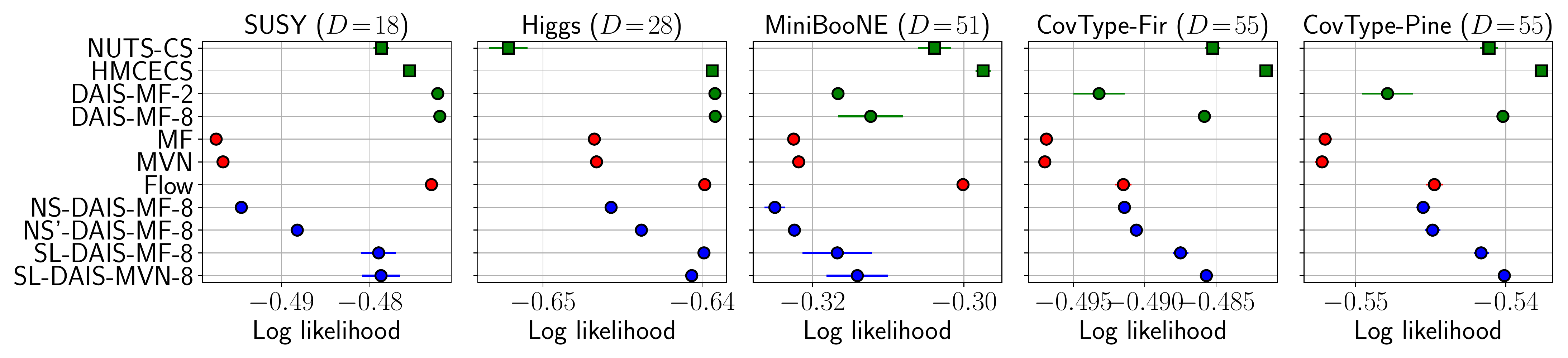}
\includegraphics[width=.95\columnwidth]{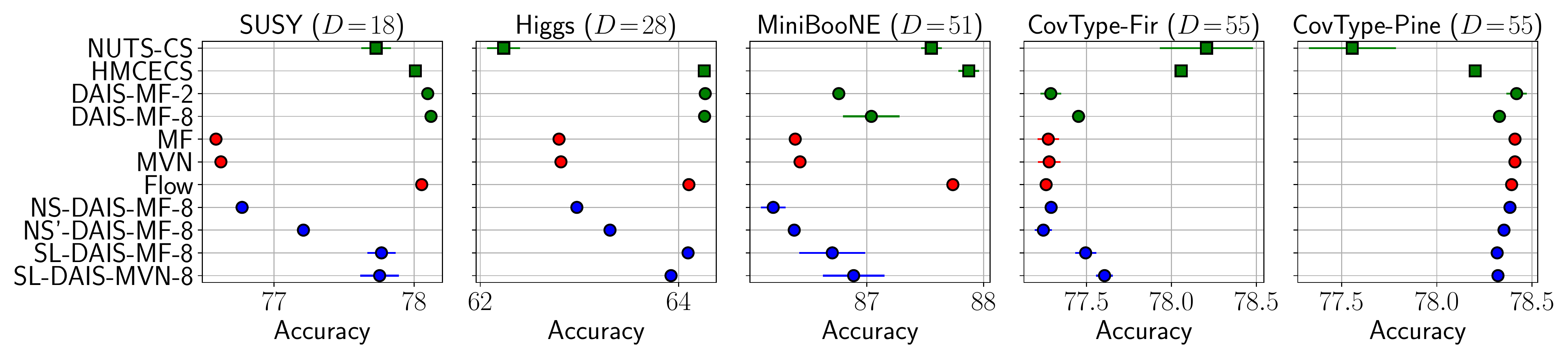}
    \caption{In this companion figure to Fig.~\ref{fig:logistic} we report full results for the logistic regression experiment in Sec.~\ref{sec:logistic}.
         Circles denote variational methods and squares denote MCMC methods.
         Metrics are averaged over 7 independent replications (except for DAIS and HMCECS where we do 3 independent replications),
         and error bars denote standard errors. Note that these figures include an additional method, namely NS$^\prime$-DAIS.
         This method is identical to NS-DAIS except that it utilizes independently drawn mini-batches of data for each evaluation
         of the annealed potential energy $V_k$ in Eqn.~\ref{eqn:vk}. We note that NS$^\prime$-DAIS outperforms NS-DAIS but is
         inferior to SL-DAIS.}
    \label{appfig:logistic}
\end{center}
\end{figure*}
%%%

%%%%%%%%%%%%%%%%%%%%%%
\subsubsection{Gaussian process classification}
\label{app:class}
%%%%%%%%%%%%%%%%%%%%%%

We consider a Gaussian process model for binary classification with a logistic link function.
The covariates are assumed to be $d$-dimensional, $\bx_n \in \RR^d$.
To make GP inference scalable we use the FITC approximation \cite{snelson2006sparse, quinonero2005unifying}, 
which results in a model density of the following form:
%%%
\begin{equation}
    \nonumber
    p(\DD, \bff, \bu) = p(\bu | \zero, \bK_{ZZ}) \prod_n p(y_n | \sigma(f_n)) p(f_n | m_n(\bu), v_n)
\end{equation}
%%%
where $\bZ \in \RR^{M\times d}$ are the $M$ inducing point locations, $\bK$ refers to the RBF kernel parameterized by
a (scalar) kernel scale and a $d$-dimensional vector of length scales, $\sigma(\cdot)$ is the sigmoid function,
$\bK_{ZZ} \in \RR^{M \times M}$ is the prior covariance, $\bu \in \RR^M$ are the inducing point values, $f_n$ is the Gaussian process function
value corresponding to covariate $\bx_n$, and $m_n(\bu)$ and $v_n$ are given by:
%%%
\begin{equation}
    \begin{split}
        m_n(\bu) = \bk_{nZ}^{\rm T} \bK_{ZZ}^{-1} \bu \qquad \qquad
        v_n = k(\bx_n, \bx_n) - \bk_{nZ}^{\rm T} \bK_{ZZ}^{-1}  \bk_{nZ}
    \end{split}
\end{equation}
%%%
Here $\bk_{nZ} \in \RR^{M}$ with $(\bk_{nZ})_m = k(\bx_n, \bz_m)$ for $m=1,...,M$, where $k(\cdot, \cdot)$ is the kernel function.
Since by construction the $\{ f_n \}$ are uncorrelated with one another, we can approximately integrate out $\{ f_n \}$ using
Gauss-Hermite quadrature using $Q$ quadrature points. In all our experiments we use $Q=16$, $M=128$, and $\Nsurr=512$.
This results in a model density of the form
%%%
\begin{equation}
    \label{eqn:fitc}
    p(\DD, \bu) = p(\bu | \zero, \bK_{ZZ}) \prod_n p(y_n | \bu, \bx_n) 
\end{equation}
%%%
where the cost of computing $N$ likelihoods $p(y_n | \bu, \bx_n)$ is $\OO(NM^2 + M^3 + NQ)$.
We use variational inference to infer an approximate posterior over $\bu$ for the model density in Eqn.~\ref{eqn:fitc}.
In practice $N \gg M$ and the cost of computing the full likelihood is dominated by
the $\OO(NM^2)$ term.
In all our Gaussian process experiments we initialize the inducing point locations $\bZ$ with k-means and treat them as learnable parameters. We use a mini-batch size $B=128$. After training ELBOs are estimated with 20k MC samples.
The MVN base distribution of SL-DAIS is initialized using the result from the MVN baseline. 

%%%%%%%%%%%%%%%%%%%%%%
\subsubsection{Robust Gaussian process regression}
\label{app:robust}
%%%%%%%%%%%%%%%%%%%%%%

We proceed exactly as in Sec.~\ref{app:class} except
we use a Student's t likelihood $p(y_n | \nu, f_n, \sigma_{\rm obs}) = {\rm StudentT}(y_n | \nu, f_n, \sigma_{\rm obs})$.
Here $\nu > 0$ is the overdispersion parameter, with small $\nu$ corresponding to large overdispersion.
We constrain $\nu$ to satisfy $\nu > 2$, which is equivalent to requiring that the observation noise have finite variance.

The Precipitation dataset is available from the IRI/LDEO Climate Data Library.\footnote{
  \url{http://iridl.ldeo.columbia.edu/maproom/Global/Precipitation/WASP_Indices.html}
}
This spatio-temporal dataset represents the `WASP' index (Weighted Anomaly Standardized Precipitation) at various latitudes and longitudes.
Each data point corresponds to the WASP index for a given year (which itself is the average of monthly WASP indices).
We use the data from 2010, resulting in a dataset with $10127$ data points.
In each train/test split we randomly choose $N=8000$ data points for training and use the remaining $2127$ data points for testing.

We use a mini-batch size $B=128$. We estimate the final ELBO using $5000$ samples
and test log likelihoods and accuracies using $1000$ posterior samples.
We use $\Nsurr=256$ surrogate data points with the \texttt{RAND} surrogate parameterization (see Sec.~\ref{app:ablation}).
We use an initial step size $\eta_{\rm init}=10^{-4}$ and initialize SL-DAIS using the model and variational parameters
obtained with a MVN variational distribution.

\begin{table*}[t]
    \centering
    \resizebox{0.85\linewidth}{!}{%
\begin{tabular}{llllllll}
\hline
\small{}         & \small{MF} & \small{MVN} & \small{Flow} & \small{NS-DAIS-MF} & \small{SL-DAIS-MF} & \small{NS-DAIS-MVN} & \small{SL-DAIS-MVN}  \\ \hline
\small{ELBO}           & \small{$6.60$} & \small{$5.36$} & \small{\bf 2.08} & \small{$5.00$} & \small{$2.60$} & \small{$4.16$} & \small{$2.20$}  \\
\small{Log likelihood} & \small{$6.16$} & \small{$5.88$} & \small{\bf 2.16} & \small{$4.60$} & \small{$2.44$} & \small{$4.36$} & \small{$2.40$} \\
\small{Accuracy}       & \small{$5.48$} & \small{$5.24$} & \small{\bf 1.96} & \small{$4.52$} & \small{$3.40$} & \small{$4.12$} & \small{$3.28$} \\ \hline
\end{tabular}
     } % end resize
\caption{We report performance ranks w.r.t.~three metrics across 5 train/test splits and 5 datasets for the logistic regression
    experiment in Sec.~\ref{sec:logistic}. Lower is better for all metrics. The rank satisfies $1 \le {\rm rank} \le 7$, since we compare 7 scalable variational methods.}
\label{table:rankfull}
\end{table*}

%%%
\begin{figure}[ht]
\begin{center}
\includegraphics[width=0.46\columnwidth]{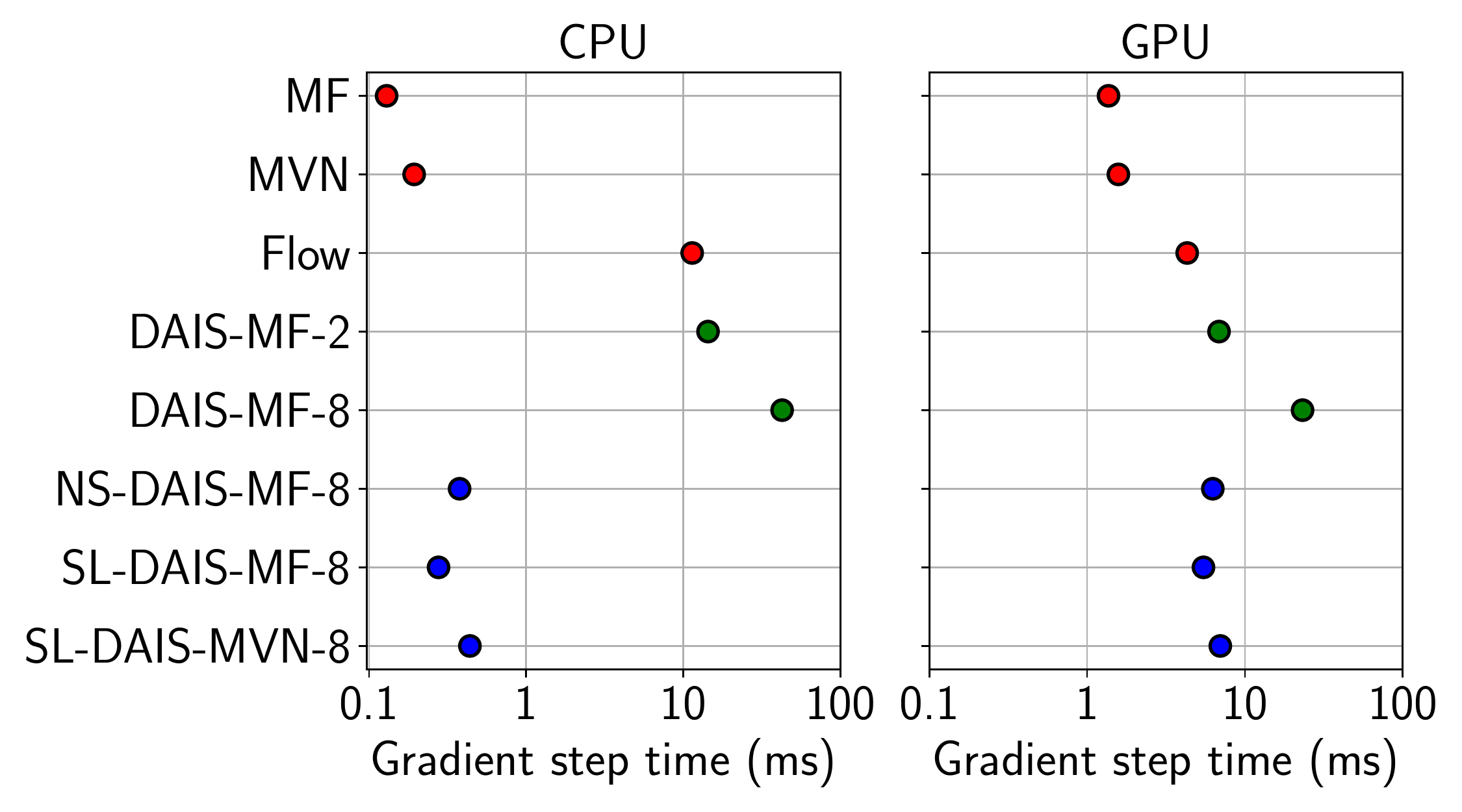}
    \caption{We report the time per optimization step for each variational method in Fig.~\ref{fig:logistic}
    on the CovType-Fir dataset.
    We compare CPU runtime (24 cores; Intel Xeon Gold 5220R 2.2GHz) to GPU runtime (NVIDIA Tesla K80).
    We note that the runtime comparison against DAIS would be increasingly favorable to NS-DAIS and SL-DAIS as the 
    size of the dataset increases.
    }
    \label{fig:timing}
\end{center}
\end{figure}
%%%

%%%%%%%%%%%%%%%%%%%%%%
\subsubsection{Local latent variable models}
\label{app:llvmexp}
%%%%%%%%%%%%%%%%%%%%%%

We use the Pol UCI dataset with 15k training and 15k test data points.
We use 10k MC samples to estimate ELBOs after training. We set $B=256$.

%%%%%%%%%%%%%%%%%%%%%%
\subsection{Additional figures and tables}
\label{app:expadd}
%%%%%%%%%%%%%%%%%%%%%%

For additional results pertaining to the logistic regression experiment in Sec.~\ref{sec:logistic} see 
Fig.~\ref{appfig:logistic}, Fig.~\ref{fig:timing}, and Table~\ref{table:rankfull}.
For an expanded version of the table in Sec.~\ref{sec:ablation} see Table~\ref{table:ablation_full}.

\end{document}